\documentclass{article}
\usepackage[utf8]{inputenc}
\usepackage{amsmath}
\usepackage{amssymb}
\usepackage{amsmath}
\usepackage{nccmath}
\usepackage{cleveref}
\usepackage{bm}
\usepackage{graphicx}
\usepackage{authblk}
\usepackage{lineno}
\usepackage{xr}
\usepackage{float}
\usepackage{url}

\usepackage[obeyFinal,textsize=footnotesize]{todonotes}

\makeatletter
\newcommand*{\addFileDependency}[1]{
  \typeout{(#1)}
  \@addtofilelist{#1}
  \IfFileExists{#1}{}{\typeout{No file #1.}}
}
\makeatother

\newcommand*\sref[1]{%
    Supplement S\ref{#1}}
 
\newcommand*\sfref[1]{%
    Supplementary Figure \ref{#1}}
 
\DeclareMathOperator{\trace}{Tr}
\DeclareMathOperator{\Hess}{Hess}

\title{Universal Differential Equations for Scientific Machine Learning}
\author[a,b]{Christopher Rackauckas}
\author[c]{Yingbo Ma}
\author[d]{Julius Martensen}
\author[a]{Collin Warner}
\author[e]{Kirill Zubov}
\author[a]{Rohit Supekar}
\author[a]{Dominic Skinner}
\author[a]{Ali Ramadhan}
\author[a]{Alan Edelman}

\affil[a]{Massachusetts Institute of Technology}
\affil[b]{University of Maryland, Baltimore}
\affil[c]{Julia Computing}
\affil[d]{Otto von Guericke University, Magdeburg}
\affil[e]{Saint Petersburg State University}

\begin{document}

\maketitle

\begin{abstract}
In the context of science, the well-known adage ``a picture is worth a thousand words'' might well be ``a model is worth a thousand datasets.'' In this manuscript we introduce the SciML software ecosystem as a tool for mixing the information of physical laws and scientific models with data-driven machine learning approaches. We describe a mathematical object, which we denote universal differential equations (UDEs), as the unifying framework connecting the ecosystem. We show how a wide variety of applications, from automatically discovering biological mechanisms to solving high-dimensional Hamilton-Jacobi-Bellman equations, can be phrased and efficiently handled through the UDE formalism and its tooling. We demonstrate the generality of the software tooling to handle stochasticity, delays, and implicit constraints. This funnels the wide variety of SciML applications into a core set of training mechanisms which are highly optimized, stabilized for stiff equations, and compatible with distributed parallelism and GPU accelerators. 
\end{abstract}

\section{Introduction}

Recent advances in machine learning have been dominated by deep learning which uses readily available ``big data'' to solve previously difficult problems such as image recognition \cite{traore2018deep,8550021,chen2019looks} and natural language processing \cite{young2018recent,otter2018survey,tsuruoka2019deep}. While some areas of science have begun to generate the large amounts of data required to train deep learning models, notably bioinformatics \cite{li2019deep,tang2019recent,zou2019primer,angermueller2016deep,bacciu2018bioinformatics}, in many areas the expense of scientific experiments has prohibited the effectiveness of these ground breaking techniques. In these domains, mechanistic models are still predominantly deployed due to the inaccuracy of deep learning techniques with small training datasets. While these mechanistic models are constrained to be predictive by uses prior structural knowledge (i.e. a form of inductive bias), the data-driven approach of machine learning can be more flexible and allows one to drop the simplifying assumptions required to derive theoretical models. The purpose of this work is to give a mathematical framework and accompanying software tool which bridges the gap by merging the best of both methodologies while mitigating the deficiencies. 

This work falls into the burgeoning field of ``scientific machine learning'' which seeks to integrate machine learning derived idea into traditional engineering-related methods in order to utilize domain knowledge and known physical information into the learning process \cite{baker2019workshop}. It has recently been shown to be advantageous to merge differential equations with machine learning. Physics-Informed Neural Networks (PINNs) use partial differential equations in the cost functions of neural networks to incorporate prior scientific knowledge \cite{raissi2019physics}. While this has been shown to be a form of data-efficient machine learning for some scientific applications, PINNs frame the solution process as a large optimization. The data efficiency of PINNs has been shown to further improve when encoding structural assumptions like energy conservation by directly modeling the Hamiltonian \cite{greydanus2019hamiltonian,zhong2019symplectic}. While this formalism incorporates the knowledge of physical systems into machine learning, it does not incorporate the numerical techniques which have led to stable and efficient solvers the large majority of scientific models. Treating the training process as fully implicit is compute-intensive which recent results have shown is increasingly difficult as the stiffness of the model increases \cite{wang2020understanding}. While work has demonstrated that efficiency is greatly improved when incorporating classical discretization techniques into the training process (as ``discrete physics-informed neural networks'', such as in multi-step neural networks \cite{raissi2018multistep}), the formalism provided by PINNs is not conducive to the full use of classical model simulation software and thus every major PINN software framework, such as deepxde \cite{lu2019deepxde} and SimNet \cite{hennigh2020nvidia}, does not have the ability to automatically combine scientific machine learning training with the highly efficient differential equation solvers and adjoint techniques developed over the last century.

Thus as a basis for scientific machine learning that incorporates the efficient numerical solver and adjoint techniques, we developed a formalism which we denote the universal differential equation (UDE). UDEs are differential equations which are defined in full or part by a universal approximator. A universal approximator is a parameterized object capable of representing any possible function in some parameter size limit. Common universal approximators in low dimensions include Fourier or Chebyshev expansions, while common universal approximators in high dimensions include neural networks \cite{lin2018resnet,winkler2017performance,gorban1998general,allan1999approximation,park2020minimum} and other models used through machine learning. Mathematically in its most general form, the UDE is a forced stochastic delay partial differential equation (PDE) defined with embedded universal approximators:
\begin{equation}
    \mathcal{N}[u(t),u(\alpha(t)),W(t),U_\theta(u,\beta(t))] = 0
\end{equation}
where $\alpha(t)$ is a delay function and $W(t)$ is the Wiener process. 

In this manuscript we describe how the tools of the SciML software ecosystem give rise to efficient training and analysis of the wide variety of UDEs which show up throughout the scientific literature. An astute reader may recognize a neural ordinary differential equation $u^\prime = \text{NN}_\theta(u,t)$ as the specific case of a one-dimensional UDE defined in full by a neural network \cite{alvarez2009latent,hu2014coupled,alvarez2010switched,chen2018neural,rubanova2019latent,kidger2020neural}. The previously mentioned discrete physics-informed neural networks also arise when one uses the appropriate fixed time step method as the solver for the UDE. In addition, neural network based optimal control \cite{ARAHAL1995239}, i.e. finding a neural network $c(u,t)$ which minimizes a loss of a differential equation solution $u^\prime = f(u,c(u,t),t)$, and model augmentation \cite{de2019neural} can similarly be phrased within this framework. Therefore, this architecture has an expansive reach in its utility throughout SciML and control applications.

\begin{figure}
	\centering
	\includegraphics[width=\linewidth]{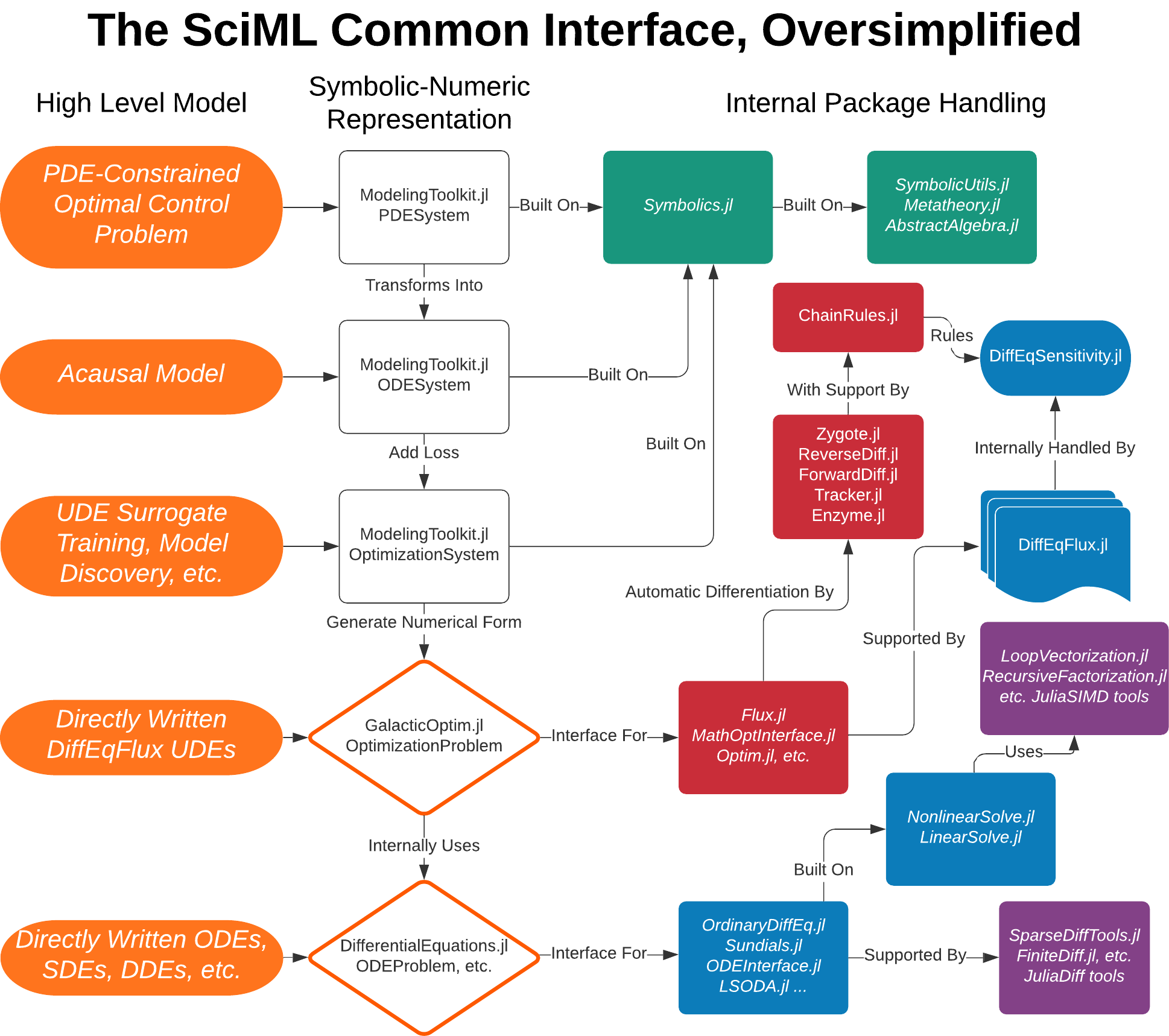}
	\caption{Simplified flow chart of the SciML software ecosystem. The orange boxes on the left denote high level use cases, from PDE-constrained optimal control and acausal modeling to direct construction of UDEs and other forms of differential equations for numerical solving. The white boxes denote the ModelingToolkit.jl symbolic-numeric system \cite{ma2021modelingtoolkit} used for automated construction of symbolically-optimized numerical code. These symbolic tools are built on the Symbolics.jl computer algebra system \cite{gowda2021high} developed and maintained by the SciML developers. The symbolic tools or direct user input generate numerical descriptions of the mathematical problems to solve, shown in the orange triangles. The OptimizationProblem structs are consumed by the GalacticOptim.jl global and local optimization library while the differential equation models of any form are consumed by DifferentialEquations.jl \cite{DifferentialEquations.jl-2017}. These models can be composed and stacked, i.e. OptimizationProblems containing ODEProblems. These tools then use underlying numerical solvers, blue denoting part of the SciML ecosystem, purple libraries denoting external libraries developed and maintained by SciML developers, red denoting external libraries used and contributed to by SciML developers.}
	\label{fig:sciml_flowchart}
\end{figure}

It is thus in this context that we demonstrate the SciML ecosystem as a set of tools for handling the wide array of possible UDEs with solvers and adjoints handling all of the cases of adaptivity, stiffness, stochasticity, delays, and more. 
Figure \ref{fig:sciml_flowchart} gives a general overview of how the tools of the SciML ecosystem connect to give rise to a modular and composable toolkit for scientific machine learning. The examples of this paper are constructed as the interplay of over 200 dependent libraries constructed by and for the SciML ecosystem, from high-level symbolic computing tools to low-level customized BLAS implementations to improve the performance of internal matrix factorizations over commonly used tools like OpenBLAS or MKL. To simplify the discussion, we will focus on how the composition of the numerical libraries and compiler-based code generation mechanisms gives rise to an efficient computing stack for SciML applications.

In Section \ref{sec:UDESoftware} we showcase how the DifferentialEquations.jl solvers, the DiffEqSensitivity.jl adjoint methods, and the DiffEqFlux.jl helper functions combine to give rise to efficient and stable training framework for UDEs. In the sections following, we focus on use cases which go beyond the obvious neural ODE and optimal control applications in order to demonstrate the generality of the framework and formalism. In Section \ref{sec:discover_ode} we highlight how the DataDrivenDiffEq.jl symbolic regression tooling can be used in conjunction with the UDE framework in order to augment models with known physical information and decrease the data requirements for symbolic model discovery. In Section \ref{sec:fbsde} we showcase how recent methods in solving high-dimensional parabolic partial differential equations can be phrased as a universal stochastic differential equation, and thus these methodologies can be extended to high order, implicit, and adaptive methods through this formulation on the SciML tools. In Section \ref{sec:closures} we demonstrate how the discovery of non-local operators for reduced order modeling, known as parameterizations in the climate modeling literature, can similarly be phrased as a UDE training problem. Together this shows how the SciML ecosystem and its UDE formalism substantially advances the ability for scientists and engineers to combine all scientific knowledge with the latest techniques of machine learning and numerical analysis to arrive at a highly efficient and flexible set of automated training techniques.

\section{Results}

\subsection{Efficient and Stable Training of Universal Differential Equations \label{sec:UDESoftware}}

Training a UDE amounts to minimizing a cost function $C(\theta)$ defined on $u_\theta(t)$, the current solution to the differential equation with respect to the choice of parameters $\theta$. One choice of cost function is the Euclidean distance $C(\theta) = \sum_i \left\Vert u_\theta(t_i) - d_i \right\Vert$ at discrete data points $(t_i,d_i)$. When optimized with local derivative-based methods, such as stochastic gradient decent, ADAM \cite{kingma2014adam}, or L-BFGS \cite{liu1989limited}, this requires the calculation of $\frac{dC}{d\theta}$ which by the chain rule amounts to calculating $\frac{du}{d\theta}$. Thus the problem of efficiently training a UDE reduces to calculating gradients of the differential equation solution with respect to parameters. 

Efficient methods for calculating these gradients are commonly called adjoints \cite{errico1997adjoint,allaire:hal-01242950,strang2007computational,hindmarsh2005sundials,johnson2012notes}. The asymptotic computational cost of these methods does not grow multiplicatively with the number of state variables and parameters like numerical or forward sensitivity approaches, and thus it has been shown empirically that adjoint methods are more efficient on large parameter models \cite{sengupta2014efficient,rackauckas2018comparison}. The DiffEqSensitivity.jl module uses the composable multiple dispatch based modular architecture of the DifferentialEquations.jl \cite{DifferentialEquations.jl-2017,rackauckas2019confederated} in order to extend all 300+ solvers for a wide variety of equations, including ordinary differential equations, stochastic differential equations, delay differential equations, differential-algebraic equations, stochastic delay differential equations, and hybrid equations which incorporate jumps and Levy processes. The full set of adjoint options available in this new software, which includes continuous adjoint methods and pure reverse-mode AD approaches, is described in \sref{SI:DiffEqFlux}. 

To the authors' knowledge, this is the first differential equation library which includes choices from all of the demonstrated categories of forward and adjoint sensitivity analysis methods. The importance of this fact is because each of these methods offers a substantial trade-off and thus all of these adjoints have different contexts in which they are the most applicable. Methods via solving ODEs and SDEs in reverse \cite{chen2018neural} are the common adjoint utilized in neural ODE software such as torchdiffeq and are O(1) in memory (and the neural SDE can utilize the virtual Brownian tree for O(1) in memory \cite{li2020scalable}), but are known to be unstable under certain conditions such as on stiff equations \cite{gholami2019anode,kim2021stiff}. Checkpointed interpolation adjoints \cite{hindmarsh2005sundials} are available which do not require stable reversibility of the ODEs while retaining a relatively low-memory implementation via checkpointing. Another stabilized adjoint technique is the continuous quadrature adjoint which trades off increased memory use to reduce the computational complexity with respect to parameters from cubic to linear. These methods are unique to the SciML ecosystem and have recently demonstrated around three orders of magnitude performance improvements for large stiff partial differential equations \cite{rackauckas2018comparison,kim2021stiff}. Section \ref{sec:benchmarks} and Section \ref{sec:climateparam} are noted cases which is not stable under the reversed adjoint but stable under the checkpointing and quadrature adjoint approach, where in the former it is demonstrated that the adjoint techniques of alternative packages diverge due to this numerical issue. All of the aforementioned adjoint methods fall under the continuous optimize-then-discretize approach. Through the integration with automatic differentiation, discrete adjoint sensitivity analysis \cite{zhang2017discrete,lauss2018discrete} is implemented through both tape-based reverse-mode \cite{RevelsLubinPapamarkou2016} and source-to-source translation \cite{innes2019zygote}, with computational trade-offs between the two approaches. The former can be faster on scalarized heterogeneous differential equations while the latter is more optimized for homogeneous vectorized functions calls like are demonstrated in neural networks and discretizations of partial differential equations. \sref{SI:adjointtradeoff} describes the literature behind continuous vs discrete adjoint approaches and showcases it as a general trade-off between performance and stability. Additionally, continuous and discrete forward mode sensitivity analysis approaches are also provided and optimized for equations with smaller numbers of parameters. Given the large number of variables involved in choosing a correct derivative calculation method, \sref{SI:decision_tree} describes a decision tree to guide users towards an appropriate choice. A compiler analysis of the user's differential equation function is provided by default to automatically choose an efficient adjoint choice.

As described in \sref{SI:BPAdjoints}, these adjoints use reverse-mode automatic differentiation for vector-transposed Jacobian products within the adjoint definitions to reduce the computational complexity. This has been shown to contribute up to two orders of magnitude in performance increases for large stiff differential equations over purely numerical vector-Jacobian product approaches which are common in other stiff ODE solver libraries with adjoints like Sundials CVODES and PETSc TS \cite{rackauckas2018comparison}. In addition, the module DiffEqFlux.jl handles compatibility with the Flux.jl neural network library so that common deep architectures, such as convolutional neural networks and recurrent neural networks, automatically uses all efficient backpropagation kernels whenever encountered in the derivative calculation of differential equations. Thus together these three tools, DifferentialEquations.jl, DiffEqSensitivity.jl, and DiffEqFlux.jl, combine to give a UDE training framework which covers the vast set of combinations to allow efficiently training each model.

\subsection{Features and Performance of the SciML Ecosystem}\label{sec:benchmarks}

We assessed the viability of alternative differential equation libraries for universal differential equation workflows by comparing the features and performance of the given libraries. Table \ref{tab:features} demonstrates that the SciML ecosystem is the only differential equation solver library with deep learning integration that supports stiff ODEs, DAEs, DDEs, stabilized adjoints, distributed and multithreaded computation. We note the importance of the stabilized adjoints in Section \ref{sec:climateparam} as many PDE discretizations with upwinding exhibit unconditional instability when reversed, and thus this is a crucial feature when training embedded neural networks in many PDE applications.  Table \ref{tab:performance} demonstrates that the SciML ecosystem exhibits more than an order of magnitude performance when solving ODEs against torchdiffeq of up to systems of 1 million equations. Because the adjoint calculation itself is a differential equation, this also corresponds to increased training times on scientific models. We note that torchdiffeq's adjoint calculation diverges on all but the first two examples due to the aforementioned stability problems. 

To reinforce this measure of performance, \sref{SI:Benchmarks} demonstrates a 100x performance difference over torchdiffeq when training the spiral neural ODE from \cite{chen2018neural,onken2020discretize}. Additionally we note that the author of the tfdiffeq library of TensorFlow has previous concluded ``speed is almost the same as the PyTorch (torchdiffeq) code base ($\pm 2\%$)''\footnote{https://github.com/titu1994/tfdiffeq/tree/v0.0.1-pre0\#caveats}. In addition, \sref{SI:Benchmarks} demonstrates a 1,600x performance advantage for the SciML ecosystem over torchsde using the geometric Brownian motion example from the torchsde documentation \cite{li2020scalable}. Given the computational burden, the mix of stiffness, and non-reversibility of the examples which follow in this paper, these results demonstrate that the SciML ecosystem is the first deep learning integrated differential equation software ecosystem that can train all of the equations necessary for the results of this paper. Note that this does not infer that our solvers will demonstrate more than an order of magnitude performance difference or more on all types of equations. For example, large non-stiff neural ODEs used in image classification tend to be dominated by large dense matrix multiplications and thus are in a different performance regime. However, these results demonstrate that on the equations generally derived from scientific models (ODEs derived from PDE semi-discretizations, heterogeneous differential equation systems, and neural networks in sufficiently small systems) that an order of magnitude or more performance difference exists over a wide set of cases.

\begin{table}
{\small
\begin{tabular}{|l||l|l|l|l|l|l|l|l|l|l|}
\hline
Feature     & Stiff      & DAEs       & SDEs       & DDEs       & Stabilized & DtO        & GPU        & Dist       & MT \\ \hline \hline
SciML       & \checkmark & \checkmark & \checkmark & \checkmark & \checkmark & \checkmark & \checkmark & \checkmark & \checkmark \\ \hline
torchdiffeq & 0          & 0          & 0          & 0          & 0          & \checkmark & \checkmark & \checkmark & \checkmark \\ \hline
torchsde    & 0          & 0          & \checkmark & 0          & 0          & \checkmark & \checkmark & \checkmark & \checkmark \\ \hline
tfdiffeq    & 0          & 0          & 0          & 0          & 0          & 0          & \checkmark & 0          & 0 \\ \hline
\end{tabular}}
\caption{Feature comparison of ML-augmented differential equation libraries. Columns correspond to support for stiff ODEs, then DAEs, SDEs, DDEs, stabilized non-reversing adjoints, discretize-then-optimize methods, distributed computing, and multi-threading.}
	\label{tab:features}
\end{table}

\begin{table}
{\small
\begin{tabular}{|l||l|l|l|l|l|l|l|l|}
\hline
\# of ODEs         & 3 & 28 & 768 & 3,072 & 12,288 & 49,152 & 196,608 & 786,432 \\ \hline \hline
SciML              & 1.0x   & 1.0x  & 1.0x & 1.0x  & 1.0x   & 1.0x   & 1.0x & 1.0x   \\ \hline
SciML DP5          & 1.0x   & 1.6x  & 2.8x & 2.7x  & 3.0x   & 3.0x   & 3.9x & 2.8x    \\ \hline
torchdiffeq dopri5 & 4,900x & 190x  & 840x & 280x  & 82x    & 31x    & 24x  & 17x    \\ \hline
\end{tabular}}
\caption{Relative time to solve for ML-augmented differential equation libraries (smaller is better). Non-stiff solver benchmarks from representative scientific systems were taken from \cite{hairerI} as described in \sref{SI:Benchmarks}. SciML stands for the optimal method choice out of the 300+ from SciML, which for the first is DP5, for the second is VCABM, and for the rest is ROCK4. 
}
	\label{tab:performance}
\end{table}

In the following sections we will demonstrate the various applications which are enabled by this combination of efficient libraries.

\subsection{Knowledge-Enhanced Symbolic Regression via UDEs\label{sec:discover_ode}}

Automatic reconstruction of models from observable data has been extensively studied. Many methods produce non-symbolic representations by learning functional representations \cite{long2017pde,raissi2018multistep} or through dynamic mode decomposition (DMD, eDMD) \cite{schmidDynamicModeDecomposition2010, williamsDataDrivenApproximationKoopman2015,liExtendedDynamicMode2017,takeishiLearningKoopmanInvariant}. Symbolic reconstruction of equations has utilized symbolic regressions which require a pre-chosen basis \cite{schaeffer2017learning,quade2016prediction}, or evolutionary methods to grow a basis \cite{cao2000evolutionary,DBLP:journals/corr/abs-1205-1986}. However, a common thread throughout much of the literature is that added domain knowledge constrains the problem to allow for more data-efficient reconstruction \cite{schaeffer2018extracting,DBLP:journals/corr/abs-1904-04058}. Here we detail how using UDEs with symbolic regression can improve the data efficiency and applicability of the techniques.

\subsubsection{Improved Identification of Nonlinear Interactions with Universal Ordinary Differential Equations}

As a motivating example, take the Lotka-Volterra system:

\begin{equation}
\begin{aligned}\label{eq:LV}
\dot{x} &= \alpha x - \beta xy,\\
\dot{y} &= \gamma x y - \delta y.
\end{aligned}
\end{equation}
Assume that a scientist has a time series of measurements dense in the prey $x$ and sparse in the predator $y$ from this system but knows the birth rate $\alpha$ of $x$ and assumes a linear decay of the population $y$. With only this information, a scientist can propose the knowledge-based UODE as:

\begin{equation}
\begin{aligned}
\dot{x} &= \alpha x + U_1(\theta, x,y)\\
\dot{y} &= -\theta_1 y + U_2(\theta, x,y)
\end{aligned}
\end{equation}
which is a system of ordinary differential equations that incorporates the known structure but leaves room for learning unknown interactions between the the predator and prey populations. Learning the unknown interactions corresponds training the UA $U: \mathbb{R}^2 \rightarrow \mathbb{R}^2$ in this UODE. \sref{SI:LV} describes the full details of the UODE training hyperparameters as well as additional examples.

\begin{figure}
	\centering
	\includegraphics[width=\linewidth]{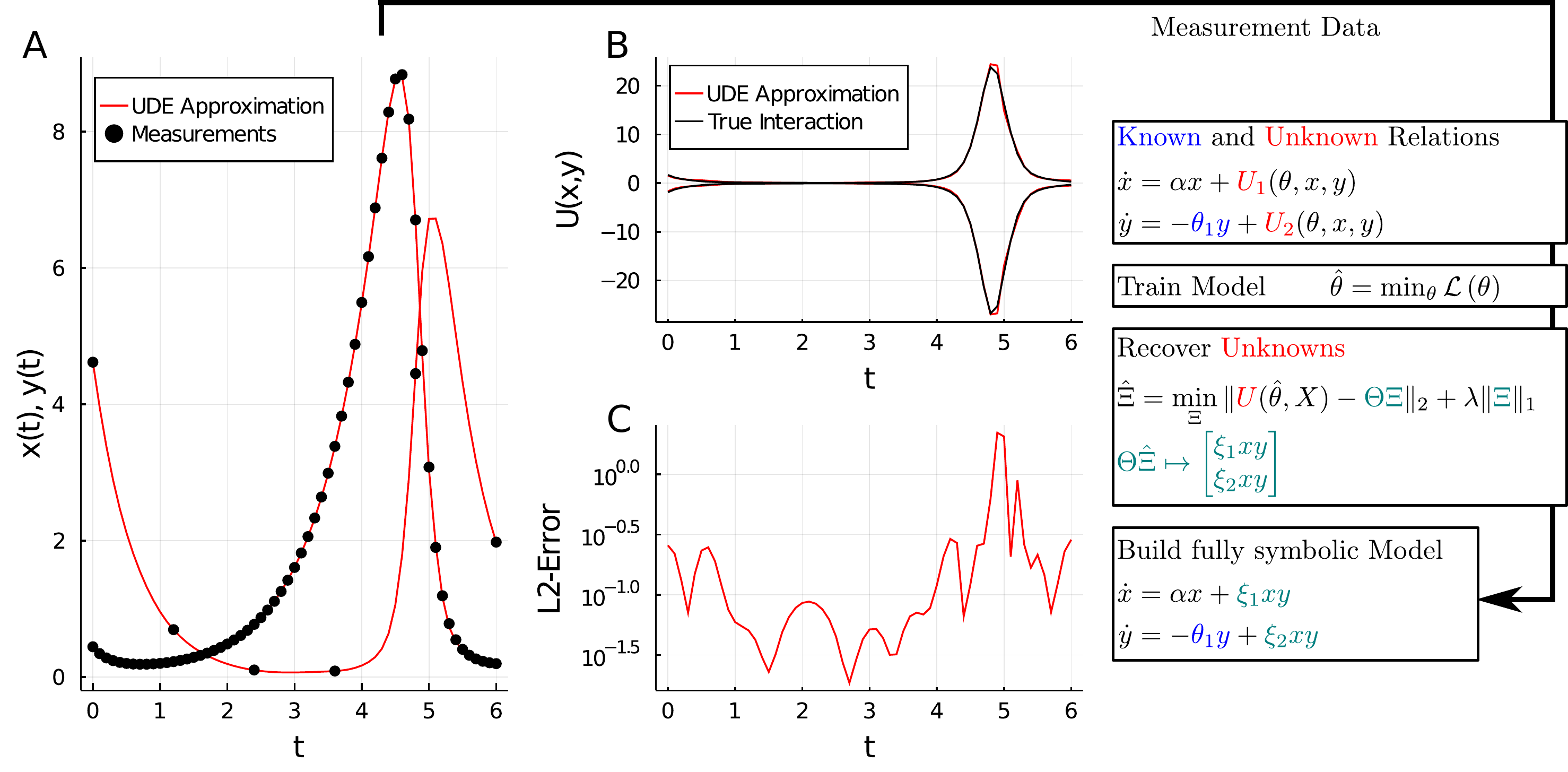}
	\caption{Automated Lotka-Volterra equation discovery with UODE-enhanced SINDy. The known and unknown parts of the model are trained against the data (A), using neural networks to reduce the complexity of the unknown equations. Afterwards symbolic regression is used to generate a fully symbolic model. (B) shows the estimation of the true polynomial interactions (black) and the neural network (red), (C) the resulting error over time.}
	\label{fig:SINDy_Training}
\end{figure}

\begin{figure}
	\centering
	\includegraphics[width=\linewidth]{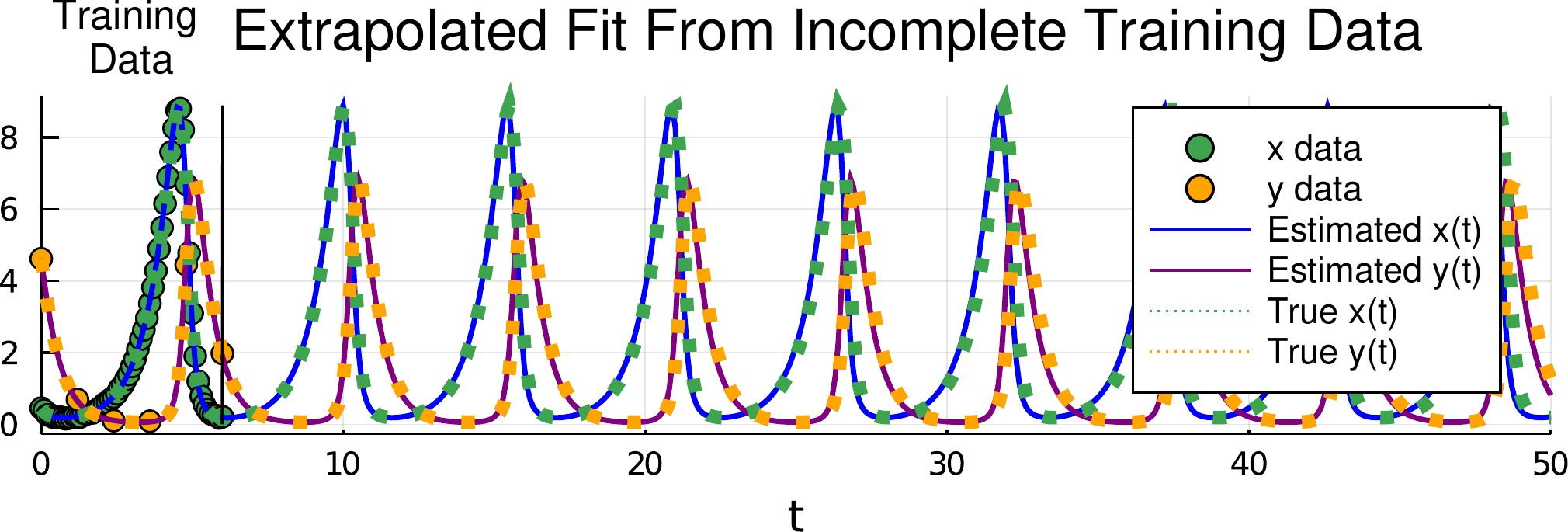}
	\caption{The extrapolation of the knowledge-enhanced SINDy fit series. The green and yellow dots show the data that was used to fit the UODE, and the dots show the true solution of the Lotka-Volterra Equations \ref{eq:LV} beyond the training data. The blue and purple lines show the extrapolated solution how the UODE-enhanced SINDy recovered equations.}
	\label{fig:SINDy_estimation}
\end{figure}

To then learn the missing portions of the equation, we can directly apply symbolic regression to the trained UAs in order to arrive at symbolic equations suggesting the missing equations. In contrast, the popular SINDy method normally \cite{bruntonDiscoveringGoverningEquations2016,manganInferringBiologicalNetworks2016,manganModelSelectionDynamical2017} approximates derivatives using a spline over the data points and subsequently applies symbolic regression to the full equation. The use of such interpolating spline techniques for derivative estimates implies dense enough data for accurate derivative calculations, which is not required in the UODE framework as the trained neural network can be sampled continuously and information can be spread throughout different data sources (such as structural inductive bias) as long as a causal and identifiable relation within the system is present.
Figure \ref{fig:SINDy_Training} shows the UODE-based symbolic regression,
Figure \ref{fig:SINDy_estimation} showcases its ability to fully and accurately recover the correct symbolic equations from the limited data, while the same symbolic regression method with the same data used in the SINDy approach with derivative smoothing is unable to recover the equations. \sref{SI:LV} further demonstrates additional examples using this approach and showcases how the UDE improves the performance in the presense of noise. Together this highlights how incorporating prior structural knowledge through the UDE framework can improve the performance of symbolic regression techniques.

\subsubsection{Generalizing Sparse Regression of Non-ODEs via UDEs}\label{sec:sparseregression}

The use of UDEs for extending sparse regression training methods also applies to extending the types of problems which sparse regression can be applied to. For example, one might wish to encode prior knowledge of the conservation equation $1 =  \sum_i u_{i}(t)$ to the evolution of the states of a discovered model. In this case, a universal differential-algebraic equation (DAE) of the form:
\begin{align}
\frac{du}{dt} &= U_\theta(u), \\
1 &=  \sum_i u_{i}(t),
\end{align}
can be utilized to encode this prior knowledge as a DAE $Mu^\prime = U_\theta(u)$ where $M$ is singular. One the UA is trained, symbolic regression on the $U_\theta(u)$ subsequently discovers the dynamical equations. Similarly, one may know the general evolution of the system but be unaware of the stochastic features, leading to a universal stochastic differential equation of the form:
\begin{align}
    du = f(u,t)dt + g(u,U_\theta(u))dW_t.
\end{align}
Tutorials in DiffEqFlux.jl showcase how to train the embedded UAs within these objects, which can then subsequently be symbolically regressed upon as in Section \ref{sec:discover_ode}.

\begin{figure}[H]
	\centering
	\includegraphics[width=\linewidth]{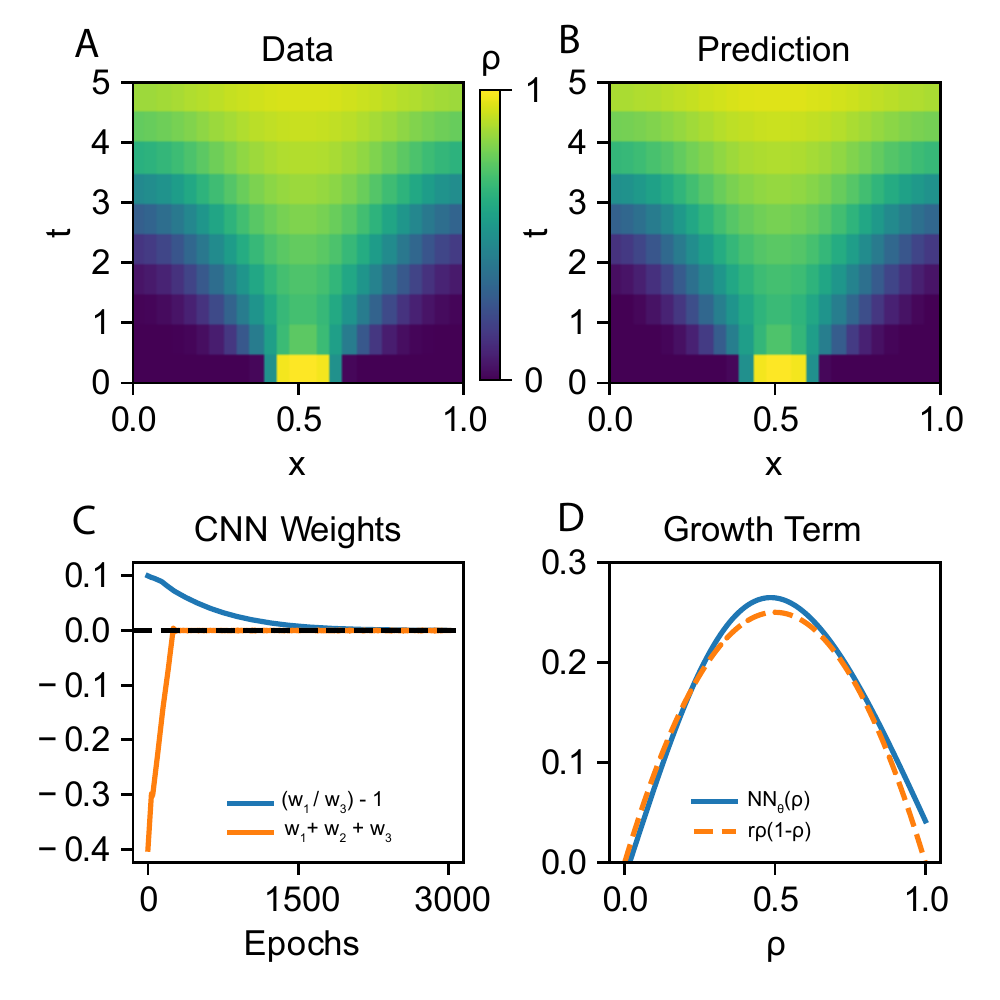}
	\caption{Recovery of the UPDE for the Fisher-KPP equation. (A) Training data and (B) prediction of the UPDE for $\rho(x,t)$. (C) Curves for the weights of the CNN filter $[w_1, w_2, w_3]$ indicate the recovery of the $[1, -2, 1]$ stencil for the 1-dimensional Laplacian. (D) Comparison of the learned (blue) and the true growth term (orange) showcases the learned parabolic form of the missing nonlinear equation.}
	\label{fig:fisher-kpp}
\end{figure}

As a concrete example, we show how UDEs can turn equation discovery of partial differential equations into a low-dimensional sparse regression problem. To demonstrate discovery of spatio-temporal equations directly from data, we consider data generated from the one-dimensional Fisher-KPP \\ (Kolmogorov–Petrovsky–Piskunov) PDE \cite{Fisher1937}:

\begin{equation}
\frac{\partial \rho}{\partial t} = r\rho(1-\rho) + D \frac{\partial^2 \rho}{\partial x^2}, \label{Eqn:Fisher-KPP}
\end{equation}
with periodic boundary conditions. Such reaction-diffusion equations appear in diverse physical, chemical and biological problems \cite{Grindrod1996}. To learn from the generated data, we define the UPDE:
\begin{equation}
\rho_t = \text{U}_\theta(\rho) + \hat{D}\,\text{CNN}(\rho),
\label{Eqn:UPDE-Fisher}
\end{equation}
and train the UAs within this equation by performing a method of lines discretization as described in \sref{SI:KPPUA}. Note that the derivative operator is approximated as a convolutional neural network $\text{CNN}$, a learnable arbitrary representation of a stencil while treating the coefficient $\hat{D}$ as an unknown parameter fit simultaneously with the neural network weights. In this representation, the nonlinear term $U_\theta$ of the semilinear partial differential equation is simply an $\mathbb{R}\rightarrow\mathbb{R}$ operator locally applied to each point in space. We note that \sref{SI:KPPUA} showcases how using the Fourier layer UA provided by DiffEqFlux.jl is an order of magnitude more efficient in training time than using a neural network for this low-dimensional case. This incorporation of prior structural information greatly decreases the dimensionality of the sparse regression as opposed to direct sparse regression techniques like PDE-FIND \cite{Rudy2017}, changing the sparse regression from acting on a $\mathcal{O}(n^d)$ state variables for $n$ discretization points in $d$ dimensions to simply $\mathcal{O}(1)$. Given the cost scaling of the sparse regression techniques is in many cases $\mathcal{O}(k^2 m + m^2 k)$ for $k$ states and $m$ data points, this amounts to a substantial decrease in the asymptotic cost of the symbolic regression by incorporating the inductive bias of the UDE. Figure \ref{fig:fisher-kpp} shows the result of training the UPDE against the simulated data, which recovers the canonical $[1,-2,1]$ stencil of the one-dimensional Laplacian and the diffusion constant, while symbolic regression of the $U_\theta$ term recovers the quadratic growth term. 


\subsection{Computationally-Efficient Solving of \\High-Dimensional Partial Differential \\Equations \label{sec:fbsde}}

It is impractical to solve high dimensional PDEs with mesh-based techniques since the number of mesh points scales exponentially with the number of dimensions. Given this difficulty, mesh-free methods based on universal approximators such as neural networks have been constructed to allow for direct solving of high dimensional PDEs \cite{Sirignano_2018,lagaris1998artificial}. Recently, methods based on transforming partial differential equations into alternative forms, such as backwards stochastic differential equations (BSDEs), which are then approximated by neural networks have been shown to be highly efficient on important equations such as the nonlinear Black-Scholes and Hamilton-Jacobi-Bellman (HJB) equations \cite{weinan2017deep,Han8505,zang2019weak,hure2019some}. Here we will showcase how one of these methods, a deep BSDE method for semilinear parabolic equations \cite{Han8505}, can be reinterpreted as a universal stochastic differential equation (USDE) to generalize the method and allow for enhancements like adaptivity, higher order integration for increased efficiency, and handling of stiff driving equations through the SciML software.

Consider the class of semilinear parabolic PDEs with a finite time span $t\in[0, T]$ and $d$-dimensional space $x\in\mathbb R^d$ that have the form:

\begin{equation}
\begin{aligned}
\label{pdeform}
\frac{\partial u}{\partial t}(t,x) 	&+\frac{1}{2}\trace\left(\sigma\sigma^{T}(t,x)\left(\Hess_{x}u\right)(t,x)\right)\\
&+\nabla u(t,x)\cdot\mu(t,x) \\
&+f\left(t,x,u(t,x),\sigma^{T}(t,x)\nabla u(t,x)\right)=0,
\end{aligned}
\end{equation}
with a terminal condition $u(T,x)=g(x)$.  \sref{SI:HJB} describes how this PDE can be solved 
by approximating by approximating the FBSDE:

\begin{equation}
\label{composite_sde2}
\begin{aligned}
dX_t &= \mu(t,X_t) dt + \sigma (t,X_t) dW_t,\\
dU_t &= f(t,X_t,U_t,U^1_{\theta_1}(t,X_t)) dt + \left[U^1_{\theta_1}(t,X_t)\right]^T dW_t,
\end{aligned}
\end{equation}
where $U^1_{\theta_1}$ and $U^2_{\theta_2}$ are UAs and the loss function is given by the requiring that the terminating condition $g(X_T) = u(X_T,W_T)$ is satisfied.

\subsubsection{Adaptive Solution of High-Dimensional Hamilton-Jacobi-Bellman Equations}

A fixed time step Euler-Maryumana discretization of this USDE gives rise to the deep BSDE method \cite{Han8505}. However, this form as a USDE generalizes the approach in a way that makes all of the methodologies of our USDE training library readily available, such as higher order methods, adaptivity, and implicit methods for stiff SDEs. As a motivating example, consider the classical linear-quadratic Gaussian (LQG) control problem in 100 dimensions:

\begin{equation}
dX_t = 2\sqrt{\lambda} c_t dt + \sqrt{2} dW_t,
\end{equation}
with $t\in [0,T]$, $X_0 = x$, and with a cost function $C(c_t) = \mathbb{E}\left[\int_0^T \Vert c_t \Vert^2 dt + g(X_t) \right]$ where $X_t$ is the state we wish to control, $\lambda$ is the strength of the
control, and $c_t$ is the control process.  Minimizing the control corresponds to solving the 100-dimensional HJB equation \cite{chang2004stochastic,6497508}:

\begin{equation}\label{eq:HJB}
\frac{\partial u}{\partial t} + \nabla^2 u - \lambda \Vert \nabla u \Vert^2 = 0
\end{equation}

\begin{figure}
	\centering
	\includegraphics[width=\linewidth]{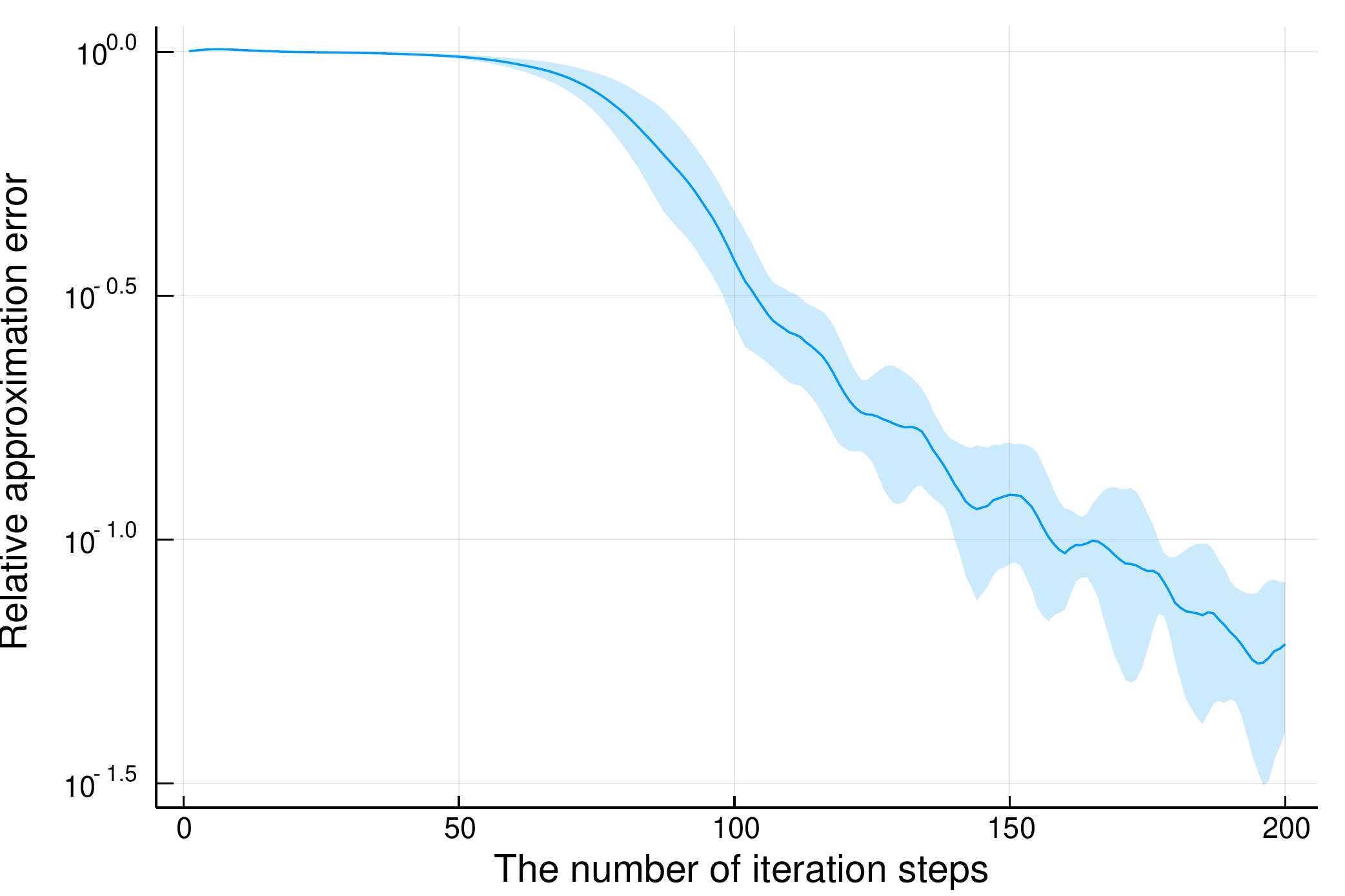}
	\caption{Adaptive solution of the 100-dimensional Hamilton-Jacobi-Bellman equation. This demonstrates that as the universal approximators $U^1_{\theta_1}$ and $U^2_{\theta_2}$ converge to satisfy the terminating condition, $U^2_{\theta_2}$ network convergences to the solution of Equation \ref{eq:HJB}.}
	\label{SIfig:hjb}
\end{figure}

We solve the PDE by training the USDE using an adaptive Euler-Maruyama method \cite{lamba2003adaptive} as described in \sref{SI:HJB}. \sfref{SIfig:hjb} showcases that this methodology accurately solves the equations, effectively extending recent algorithmic advancements to adaptive forms simply be reinterpreting the equation as a USDE. While classical methods would require an amount of memory that is exponential in the number of dimensions making classical adaptively approaches infeasible, this approach is the first the authors are aware of to generalize the high order, adaptive, highly stable software tooling to the high-dimensional PDE setting.

\subsection{Accelerated Scientific Simulation with Automatically Constructed Closure Relations}\label{sec:closures}

\subsubsection{Automated Discovery of Large-Eddy Model Parameterizations \label{sec:climateparam}}

As an example of directly accelerating existing scientific workflows, we focus on the Boussinesq equations \cite{CUSHMANROISIN201199}. The Boussinesq equations are a system of 3+1-dimensional partial differential equations acquired through simplifying assumptions on the incompressible Navier-Stokes equations, represented by the system:

\begin{equation}
\begin{aligned}
\nabla \cdot {\bf u} &= 0, \\
\frac{\partial {\bf u}}{\partial t} + ({\bf u} \cdot \nabla){\bf u} &= -\nabla p + \nu \nabla^2 {\bf u} + b\hat{z}, \\
\frac{\partial T}{\partial t} + {\bf u} \cdot \nabla T &= \kappa \nabla^2 T,
\end{aligned}
\end{equation}
where ${\bf u} = (u,v,w)$ is the fluid velocity, $p$ is the kinematic pressure, $\nu$ is the kinematic viscosity, $\kappa$ is the thermal diffusivity, $T$ is the temperature, and $b$ is the fluid buoyancy. We assume that density and temperature are related by a linear equation of state so that the buoyancy $b$ is only a function $b = \alpha g T$ where $\alpha$ is the thermal expansion coefficient and $g$ is the acceleration due to gravity.

This system is commonly used in climate modeling, especially as the voxels for modeling the ocean \cite{ZHANG2015347,2000291,griffies2008formulating,CUSHMANROISIN201199} in a multi-scale model that approximates these equations by averaging out the horizontal dynamics $\overline{T}(z,t) = \iint T(x,y,z,t) \, dx \, dy$ in individual boxes. The resulting approximation is a local advection-diffusion equation describing the evolution of the horizontally-averaged temperature $\overline{T}$:

\begin{equation}
\frac{\partial \overline{T}}{\partial t} + \frac{\partial \overline{wT}}{\partial z} = \kappa \frac{\partial^2 \overline{T}}{\partial z^2}.
\end{equation}
This one-dimensional approximating system is not closed since $\overline{wT}$ (the horizontal average temperature flux in the vertical direction) is unknown. Common practice closes the system by manually determining an approximating $\overline{wT}$ from ad-hoc models, physical reasoning, and scaling laws. However, we can utilize a UDE-automated approach to learn such an approximation from data. Let 
\begin{equation}
\overline{wT} = U_\theta\left(P,\overline{T},\frac{\partial \overline{T}}{\partial z}\right)
\end{equation}
where $P$ are the physical parameters of the Boussinesq equation at different regimes of the ocean, such as the amount of surface heating or the strength of the surface winds \cite{CVMix}. Using data from average temperatures $\overline{T}$ and known physical parameters $P$, the non-locality of the convection term may be captured by training a universal diffusion-advection partial differential equation. \sfref{SIfig:climate} demonstrates the accuracy of the approach using a deep UPDE with high order stabilized-explicit Runge-Kutta (ROCK) methods where the fitting is described in \sref{SI:Climate}. To contrast the trained UPDE, we directly simulated the 3D Boussinesq equations under similar physical conditions and demonstrated that the neural parameterization results in around a 15,000x acceleration. This demonstrates that physical-dependent parameterizations for acceleration can be directly learned from data utilizing the previous knowledge of the averaging approximation and mixed with a data-driven discovery approach.

\subsubsection{Data-Driven Nonlinear Closure Relations for Model Reduction in Non-Newtonian Viscoelastic Fluids \label{sec:fluids}}

\begin{figure}
	\centering
	\includegraphics[width=\linewidth]{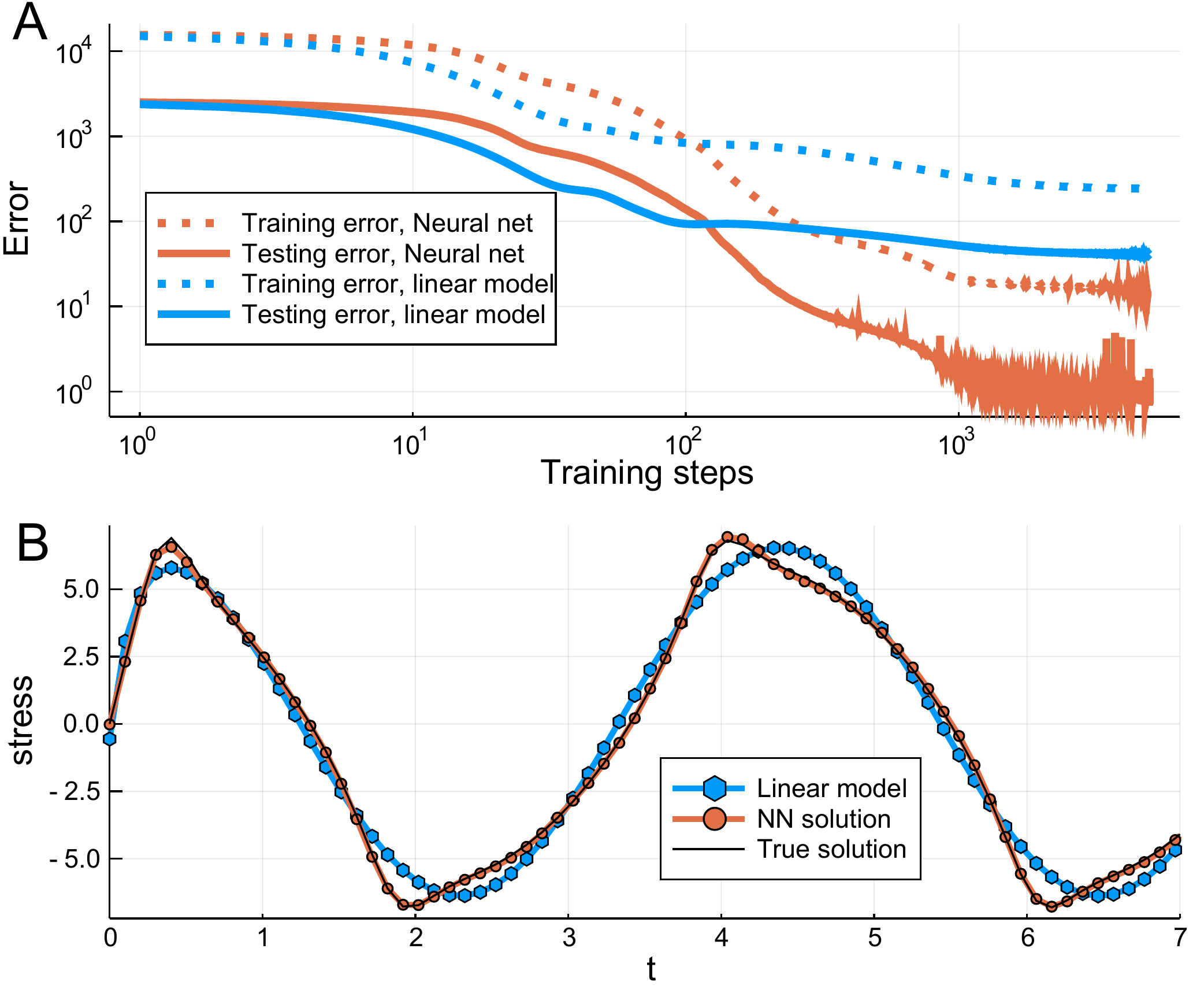}
	\caption{Convergence of neural closure relations for a non-Newtonian Fluid. (A) Error between the approximated $\sigma$ using the linear approximation Equation \ref{eq:OldroydB} and the neural network closure relation Equation \ref{eq:nnsigma}  against the full FENE-P solution. The error is measured for the strain rates $\dot{\gamma} = 12\cos \omega t$ for $\omega = 1,1.2,\ldots,2$ and tested with the strain rate $\dot{\gamma} = 12\cos 1.5 t$. (B) Predictions of stress for testing strain rate for the linear approximation and UODE solution against the exact FENE-P stress.}
	\label{fig:fluid}
\end{figure}

All continuum materials satisfy conservation equations for mass and momentum. The difference between an elastic solid and a viscous fluid comes down to the constitutive law relating the stresses and strains. In a one-dimensional system, an elastic solid satisfies $\sigma = G \gamma$, with  stress $\sigma$, strain $\gamma$, and elastic modulus $G$, whereas a viscous fluid satisfies $\sigma=\eta \dot{\gamma}$, with viscosity $\eta$ and strain rate $\dot{\gamma}$. Non-Newtonian fluids have more complex constitutive laws, for instance when stress depends on the history of deformation,
\begin{equation}
    \sigma(t) = \int_{-\infty}^t G(t-s) F(\dot{\gamma}(s)) \, \mathrm{d} s,
\end{equation}
alternatively expressed in the instantaneous form \cite{MorrisonRheology}:
%

\begin{equation}
\begin{aligned}
\sigma(t) &= \phi_1(t),\\
\frac{\mathrm{d}\phi_1}{\mathrm{d}t} &= G(0)F(\dot{\gamma}) + \phi_2,\\
\frac{\mathrm{d}\phi_2}{\mathrm{d}t} &= \frac{\mathrm{d}G(0)}{\mathrm{d}t}F(\dot{\gamma}) + \phi_3,\\
&\vdots
\end{aligned}
\end{equation}
where the history is stored in $\phi_i$. To become computationally feasible, the expansion is truncated, often in an ad-hoc manner, e.g. $\phi_n = \phi_{n+1} = \cdots = 0$, for some $n$. Only with a simple choice of $G(t)$ does an exact closure condition exist, e.g. the Oldroyd-B model. For a fully nonlinear approximation, we train a UODE according to the details in \sref{SI:Fluid} to learn a closure relation:

\begin{align}\label{eq:nnsigma}
    \sigma(t) &= U_0(\dot{\gamma},\phi_1,\ldots,\phi_N),\\
    \frac{d\phi_i}{dt} &= U_i(\dot{\gamma},\phi_1,\ldots,\phi_N),\quad \text{for $i=1$ to $N$}
\end{align}
from the numerical solution of the FENE-P equations, a fully non-linear constitutive law requiring a truncation condition \cite{Oliveira2009}. Figure \ref{fig:fluid} compares the neural network approach to a linear, Oldroyd-B like, model for $\sigma$ and showcases that the nonlinear approximation improves the accuracy by more than 50x. We note that the neural network approximation accelerates the solution by 2x over the original 6-state DAE, demonstrating that the universal differential equation approach to model acceleration is not just applicable to large-scale dynamical systems like PDEs but also can be effectively employed to accelerate small scale systems.
 
\section{Discussion}

While many attribute the success of deep learning to its blackbox nature, the key advances in deep learning applications have come from incorporating inductive bias into architectures. Deep convolutional neural networks for image processing directly utilized the local spatial structure of images by modeling convolution stencil operations. Similarly, recurrent neural networks encode a forward time progression into a deep learning model and have excelled in natural language processing and time series prediction. Here we present a software designed for generating such inductive biased architectures for a wide variety of scientific domains. Our results show that by building these hybrid mechanistic models with machine learning, we can arrive at similar efficiency advancements by utilizing all known prior knowledge of the underlying problem's structure.  While we demonstrate the utility of UDEs in equation discovery, we have also demonstrated that these methods are capable of solving many other problems such as high dimensional partial differential equations. Many methods of recent interest, such as discrete physics-informed neural networks, can additionally be written as discretizations of UDEs and thus efficiently implemented using the SciML tools.

Our software implementation is the first deep learning integrated differential equation library to include the full spectrum of adjoint sensitivity analysis methods that is required to both efficiently and accurately handle the range of training problems that can arise from universal differential equations. We have demonstrated orders of magnitude performance advantages over previous machine learning enhanced adjoint sensitivity ODE software in a variety of scientific models and demonstrated generalizations to stiff equations, DAEs, SDEs, and more. While the results of this paper span many scientific disciplines and incorporate many different modeling approaches, together all of the examples shown in this manuscript can be implemented using the SciML software ecosystem in just hundreds of lines of code each, with none of the examples taking more than half an hour to train on a standard laptop. This both demonstrates the efficiency of the software and its methodologies, along with the potential to scale to much larger applications.

\section{Code and Data Availability}

The code for reproducing the computational experiments can be found at:

\begin{verbatim}
    https://github.com/ChrisRackauckas/universal_differential_equations
\end{verbatim}

The following are the core libraries of the SciML ecosystem which together implement the functionality described in the paper:

\begin{verbatim}
    UDE training schemes: https://github.com/SciML/DiffEqFlux.jl
    Forward and adjoint rules: https://github.com/SciML/DiffEqSensitivity.jl
    Differential equation solvers: https://github.com/SciML/DifferentialEquations.jl
    Symbolic regression: https://github.com/SciML/DataDrivenDiffEq.jl
\end{verbatim}

All of the data for the experiments are simulated in the example codes.

\section{Acknowledgements}
 
We thank Jesse Bettencourt, Mike Innes, and Lyndon White for being instrumental in the early development of the DiffEqFlux.jl library, Tim Besard and Valentin Churavy for the help with the GPU tooling, and David Widmann and Kanav Gupta for their fundamental work across DifferentialEquations.jl. Special thanks to Viral Shah and Steven Johnson who have been helpful in the refinement of these ideas. We thank Charlie Strauss, Steven Johnson, Nathan Urban, and Adam Gerlach for enlightening discussions and remarks on our manuscript and software. We thank Stuart Rogers for his careful read and corrections. We thank David Duvenaud for extended discussions on this work. We thank the author of the torchsde library, Xuechen Li, for optimizing the SDE benchmark code.

\bibliographystyle{unsrt}
\bibliography{bibliography}

\pagebreak

{\Huge Supplementary Information}

\section{DiffEqFlux.jl Pullback Construction \label{SI:DiffEqFlux}}

The DiffEqFlux.jl pullback construction is not based on just one method but instead has a dispatch-based mechanism for choosing between different adjoint implementations. At a high level, the library defines the pullback on the differential equation solve function, and thus using a differential equation inside of a larger program leads to this chunk as being a single differentiable primitive that is inserted into the back pass of Flux.jl when encountered by overloading the Zygote.jl \cite{innes2019zygote} and ChainRules.jl rule sets. For any ChainRules.jl-compliant reverse-mode AD package in the Julia language, when a differential equation solve is encountered in any Julia library during the backwards pass, the adjoint method is automatically swapped in to be used for the backpropagation of the solver. The choice of the adjoint is chosen by the type of the sensealg keyword argument which are fully described below.

\subsection{Pullbacks and Adjoints}

Given a function $f(x)=y$, the pullback at $x$ is the function:

\begin{equation}
	B_f^x(y) = y^T f'(x),
\end{equation}
where $f'(x)$ is the Jacobian $J$. We note that $B_f^x(1) = \left(\nabla f\right)^T$ for a function $f$ producing a scalar output, meaning the pullback of a cost function computes the gradient. A general computer program can be written as the composition of discrete steps:

\begin{equation}
	f = f^L \circ f^{L-1} \circ \ldots \circ f^1,
\end{equation}
and thus the vector-Jacobian product can be decomposed:

\begin{equation}
	v^T J = (\ldots ((v^T J_L) J_{L-1}) \ldots ) J_1,
\end{equation}
which allows for recursively decomposing a the pullback to a primitively known set of $\mathcal{B}_{f^i}^x$:

\begin{equation}\label{eq:OldroydB}
	\mathcal{B}_{f}^{x}(A)=\mathcal{B}_{f^{1}}^{x}\left(\ldots\left(\mathcal{\mathcal{B}}_{f^{L-1}}^{x_{L-2}}\left(\mathcal{B}_{f^{L}}^{x_{L-1}}(A)\right)\right)\ldots\right),
\end{equation}
where $x_i = \left(f^{i} \circ f^{i-1} \circ \ldots \circ f^{1}\right)(x)$. Implementations of code generation for the backwards pass of an arbitrary program in a dynamic programming language can vary. For example, building a list of function compositions (a tape) is provided by libraries such as Tracker.jl \cite{Flux.jl-2018} and PyTorch \cite{NEURIPS2019_9015}, while other libraries perform direct generation of backward pass source code such as Zygote.jl \cite{innes2019zygote}, TAF \cite{giering2005generating}, and Tapenade \cite{hascoet2013tapenade}.

\subsection{Backpropagation-Accelerated DAE Adjoints for Index-1 DAEs with Constraint Equations \label{SI:BPAdjoints}}

Before describing the modes, we first describe the adjoint of the differential equation with constraints. The following derivation is based on \cite{cao2003adjoint} but modified to specialize on index 1 DAEs with linear mass matrices. The constrained ordinary differential equation:
\begin{align}
	u^\prime &= \tilde{f}(u,p,t), \\
	0 &= c(u,p,t),
\end{align}
can be rewritten in mass matrix form:
\begin{equation}
	Mu^\prime = f(u,p,t).    
\end{equation}
We wish to solve for some cost function $G(u,p)$ evaluated throughout the differential equation, i.e.:
\begin{equation}
	G(u,p) = \int_{t_0}^T g(u,p,t)dt,
\end{equation}
To derive this adjoint, introduce the Lagrange multiplier $\lambda$ to form:

\begin{equation}
	I(p) = G(p) - \int_{t_0}^T \lambda^\ast (M u^\prime - f(u,p,t))dt,
\end{equation}
Since $u^\prime = f(u,p,t)$, we have that:

\begin{equation}
	\frac{dG}{dp} = \frac{dI}{dp} = \int_{t_0}^T (g_p + g_u s)dt - \int_{t_0}^T \lambda^\ast (Ms^\prime - f_u s - f_p)dt,
\end{equation}
for $s_i$ being the sensitivity of the ith variable. After applying integration
by parts to $\lambda^\ast M s^\prime$, we require that:

\begin{align}
	M^\ast \lambda^\prime &= -\frac{df}{du}^\ast \lambda - \left(\frac{dg}{du} \right)^\ast,\\
	\lambda(T) &= 0,
\end{align}
to zero out a term and get:
\begin{equation}
	\frac{dG}{dp} = \lambda^\ast(t_0)M\frac{du}{dp}(t_0) + \int_{t_0}^T \left(g_p + \lambda^\ast f_p \right)dt.
\end{equation}
If $G$ is discrete, then it can be represented via the Dirac delta:
\begin{equation}
	G(u,p) = \int_{t_0}^T \sum_{i=1}^N \Vert d_i - u(t_i,p)\Vert^2 \delta(t_i - t)dt,
\end{equation}
in which case
\begin{equation}
	g_u(t_i) = 2(d_i - u(t_i,p)),
\end{equation}
at the data points $(t_i,d_i)$. Therefore, the derivative of an ODE solution with respect to a cost function is given by solving for $\lambda^\ast$ using an ODE for $\lambda^T$ in reverse time, and then using that to calculate $\frac{dG}{dp}$. At each time point where discrete data is found, $\lambda$ is then changed using a callback (discrete event handling) by the amount $g_u$ to represent the Dirac delta portion of the integral. Lastly, we note that $\frac{dG}{du_0} = -\lambda(0)$ in this formulation.

We have to take care of consistent initialization in the case of semi-explicit index-1 DAEs. We need to satisfy the system of equations
\begin{align}
	M^\ast \Delta \lambda^d &= h_{u^d}^\ast \lambda^a + g_{u^d}^\ast \\
	0 &= h_{u^a}^\ast \lambda^a + g_{u^a}^\ast,
\end{align}
where $^d$ and $^a$ denote differential and algebraic variables, and $f$ and $g$ denote differential and algebraic equations respectively. Combining the above two equations, we know that we need to increment the differential part of $\lambda$ by
\begin{equation}
	-h_{u^d}^\ast \left(h_{u^a}^\ast\right)^{-1} g_{u^a}^\ast + g_{u^d}^\ast
\end{equation}
at each callback. Additionally, the ODEs
\begin{equation}
	\mu' = -\lambda^\ast \frac{\partial f}{\partial p}
\end{equation}
with $\mu(T)=0$ can be appended to the system of equations to perform the quadrature for $\frac{dG}{dp}$. We note that this formulation allows for a single linear solve to generate a guaranteed consistent initialization via a linear solve. This is a major improvement over the previous technique when applied to index 1 DAEs in mass matrix form since the alternative requires a full consistent initialization of the DAE, a problem which is known to have many common failure modes \cite{kroner1997getting}. 

\subsection{Current Adjoint Calculation Methods \label{SI:adjoints}}

From this setup we have the following 8 possible modes for calculating the adjoint, with their pros and cons.

\begin{enumerate}
	\item QuadratureAdjoint: a quadrature-based approach. This utilizes interpolation of the forward solution provided by DifferentialEquations.jl to calculate $u(t)$ at arbitrary time points for doing the calculations with respect to $\frac{df}{du}$ in the reverse ODE of $\lambda$. From this a continuous interpolatable $\lambda(t)$ is generated, and the integral formula for $\frac{dG}{dp}$ is calculated using the QuadGK.jl implementation of Gauss-Kronrod quadrature. While this approach is memory heavy due to requiring the interpolation of the forward and reverse passes, it can be the fastest version for cases where the number of ODE/DAE states is small and the number of parameters is large since the QuadGK quadrature can converge faster than ODE/DAE-based versions of quadrature. This method requires an ODE or a DAE.
	\item InterpolatingAdjoint: a checkpointed interpolation approach. This approach solves the $\lambda(t)$ ODE in reverse using an interpolation of $u(t)$, but appends the equations for $\mu(t)$ and thus does not require saving the timeseries trajectory of $\lambda(t)$. For checkpointing, a scheme similar to that found in SUNDIALS \cite{hindmarsh2005sundials} is used. Points $(u_k,t_k)$ from the forward solution are chosen as the interval points. Whenever the backwards pass enters a new interval, the ODE is re-solved on $t \in [t_{k-1},t_k]$ with a continuous interpolation provided by DifferentialEquations.jl. For the reverse pass, the tstops argument is set for each $t_k$, ensuring that no backwards integration step lies in two checkpointing intervals. This requires at most a total of two forward solutions of the ODE and the memory required to hold the interpolation of the solution between two consecutive checkpoints. Note that making the checkpoints at the start and end of the integration interval makes this equivalent to a non-checkpointed interpolated approach which replaces the quadrature with an ODE/SDE/DAE solve for memory efficiency. This method tends to be both stable and require a minimal amount of memory, and is thus the default. This method requires an ODE, SDE, or a DAE.
	\item BacksolveAdjoint: a checkpointed backwards solution approach. Following \cite{chen2018neural}, after a forward solution, this approach solves the $u(t)$ equation in reverse along with the $\lambda(t)$ and $\mu(t)$ ODEs. Thus, since no interpolations are required, it requires $\mathcal{O}(1)$ memory. Unfortunately, many theoretical results show that backwards solution of ODEs is not guaranteed to be stable, and testing this adjoint on the universal partial differential equations like the diffusion-advection example of this paper showcases that it can be divergent and is thus not universally applicable, especially in cases of stiffness. Thus for stability we modify this approach by allowing checkpoints $(u_k,t_k)$ at which the reverse integration is reset, i.e. $u(t_k) = u_k$, and the backsolve is then continued. The tstops argument is set in the integrator to require that each checkpoint is hit exactly for this resetting to occur. By doing so, the resulting method utilizes $\mathcal{O}(1)$ memory + the number of checkpoints required for stability, making it take the least memory approach. However, the potential divergence does lead to small errors in the gradient, and thus for highly stiff equations we have found that this is only applicable to a certain degree of tolerance like $10^{-6}$ given reasonable numbers of checkpoints. When applicable this can be the most efficient method for large memory problems. This method requires an ODE, SDE, or a DAE.
	\item ForwardSensitivity: a forward sensitivity approach. From $u'=f(u,p,t)$, the chain rule gives $\frac{d}{dt} \frac{du}{dp} = \frac{df}{du}\frac{du}{dp} + \frac{df}{dp}$ which can be appended to the original equations to give $\frac{du}{dp}$ as a time series, which can then be used to compute $\frac{dG}{dp}$. While the computational cost of the adjoint methods scales like $\mathcal{O}(N+P)$ for $N$ differential equations and $P$ parameters, this approach scales like $\mathcal{O}(NP)$ and is thus only applicable to models with small numbers of parameters (thus excluding neural networks). However, when the universal approximator has small numbers of parameters, this can be the most efficient approach. This method requires an ODE or a DAE.
	\item ForwardDiffSensitivity: a forward-mode automatic differentiation approach, using ForwardDiff.jl \cite{RevelsLubinPapamarkou2016} to calculate the forward sensitivity equations, i.e. an AD-generated implementation of forward-mode ``discretize-then-optimize''. Because it utilizes a forward-mode approach, the scaling matches that of the forward sensitivity approach and it tends to have similar performance characteristics. This method applies to any Julia-based differential equation solver.
	\item TrackerAdjoint: a Tracker-driven taped-based reverse-mode discrete adjoint sensitivity, i.e. an AD-generated implementation of reverse-mode ``discretize-then-optimize''. This is done by using the TrackedArray constructs of Tracker.jl \cite{Flux.jl-2018} to build a Wengert list (or tape) of the forward execution of the ODE solver which is then reversed. This method applies to any Julia-based differential equation solver.
	\item ZygoteAdjoint: a Zygote-driven source-to-source reverse-mode discrete adjoint sensitivity, i.e. an AD-generated implementation of reverse-mode ``discretize-then-optimize''. This utilizes the Zygote.jl \cite{innes2019zygote} system directly on the differential equation solvers to generate a source code for the reverse pass of the solver itself. Currently this is only directly applicable to a few differential equation solvers, but is under heavy development.
	\item ReverseDiffAdjoint: A ReverseDiff.jl taped-based reverse-mode discrete adjoint sensitivity, i.e. an AD-generated implementation of reverse-mode ``discretize-then-optimize''. In contrast to TrackerAdjoint, this methodology can be substantially faster due to its ability to precompile the tape but only supports calculations on the CPU.
\end{enumerate}

For each of the non-AD approaches, there are the following choices for how the Jacobian-vector products $Jv$ (jvp) of the forward sensitivity equations and the vector-Jacobian products $v'J$ (vjp) of the adjoint sensitivity equations are computed:

\begin{enumerate}
	\item Automatic differentiation for the jvp and vjp. In this approach, automatic differentiation is utilized for directly calculating the jvps and vjps. ForwardDiff.jl with a single dual dimension is applied at $f(u+\lambda\epsilon)$ to calculate $\frac{df}{du} \lambda$ where $\epsilon$ is a dual dimensional signifier. For the vector-Jacobian products, a forward pass at $f(u)$ is utilized and the backwards pass is seeded at $\lambda$ to compute the $\lambda'\frac{df}{du}$ (and similarly for $\frac{df}{dp}$). Note that if $f$ is a neural network, this implies that this product is computed by starting the backpropagation of the neural network with $\lambda$ and the vjp is the resulting return. Four methods are allowed to be chosen for performing the internal vjp calculations:
	\begin{enumerate}
		\item Zygote.jl source-to-source transformation based vjps. Note that only non-mutating differential equation function definitions are supported in this mode. This mode is the most efficient in the presence of neural networks.
		\item Enzyme.jl source-to-source transformation basd vjps. This is the fastest vjp choice in the presence of heavy scalar operations like in chemical reaction networks, but is currently not compatible with garbage collection and thus requires non-allocating $f$ functions.
		\item ReverseDiff.jl tape-based vjps. This allows for JIT-compilation of the tape for accelerated computation. This is a the fast vjp choice in the presence of heavy scalar operations like in chemical reaction networks but more general in application than Enzyme. It is not compatible with GPU acceleration.
		\item Tracker.jl with arrays of tracked real values is utilized on mutating functions.
	\end{enumerate}
	The internal calculation of the vjp on a general UDE recurses down to primitives and embeds optimized backpropagations of the internal neural networks (and other universal approximators) for the calculation of this product when this option is used.
	\item Numerical differentiation for the jvp and vjp. In this approach, finite differences is utilized for directly calculating the jvps and vjps. For a small but finite $\epsilon$, $\left(f(u+\lambda\epsilon) - f(u)\right) / \epsilon$ is used to approximate $\frac{df}{du} \lambda$. For vjps, a finite difference gradient of $\lambda'f(u)$ is used.
	\item Automatic differentiation for Jacobian construction. In this approach, (sparse) forward-mode automatic differentiation is utilized by a combination of ForwardDiff.jl \cite{RevelsLubinPapamarkou2016} with SparseDiffTools.jl for color-vector based sparse Jacobian construction. After forming the Jacobian, the jvp or vjp is calculated.
	\item Numerical differentiation for Jacobian construction. In this approach, (sparse) numerical differentiation is utilized by a combination of DiffEqDiffTools.jl with SparseDiffTools.jl for color-vector based sparse Jacobian construction. After forming the Jacobian, the jvp or vjp is calculated.
\end{enumerate}

In total this gives 48 different adjoint method approaches, each with different performance characteristics and limitations. A full performance analysis which measures the optimal adjoint approach for various UDEs has been omitted from this paper, since the combinatorial nature of the options requires a considerable amount of space to showcase the performance advantages and generality disadvantages between each of the approaches. A follow-up study focusing on accurate performance measurements of the adjoint choice combinations on families of UDEs is planned.

\section{Sensitivity Algorithm Decision Tree}\label{SI:decision_tree}

If no \verb sensealg  is provided by the user, DiffEqSensitivity.jl will automatically choose an adjoint technique from the list. By default, a stable adjoint with an auto-adapting vjp choice is used. The following decision tree is similar to that within the defaults (but is continuously updating due to various performance changes).

\begin{itemize}
	\item If there are 50 parameters+states or less, consider using forward-mode sensititivites. If the f function is not ForwardDiff-compatible, use ForwardSensitivty, otherwise use ForwardDiffSensitivty as its more efficient.
	\item For larger equations, give BacksolveAdjoint and InterpolatingAdjoint a try. If the gradient of BacksolveAdjoint is correct, many times it's the faster choice so choose that (but it's not always faster!). If your equation is stiff or a DAE, skip this step as BacksolveAdjoint is almost certainly unstable.
	\item If your equation does not use much memory and you're using a stiff solver, consider using QuadratureAdjoint as it is asymptotically more computationally efficient by trading off memory cost.
	\item If the other methods are all unstable (check the gradients against each other!), then ReverseDiffAdjoint is a good fallback on CPU, while TrackerAdjoint is a good fallback on GPUs.
	\item After choosing a general sensealg, if the choice is InterpolatingAdjoint, QuadratureAdjoint, or BacksolveAdjoint, then optimize the choice of vjp calculation next:
	\begin{itemize}
		\item If your function has no branching (no if statements) and is heavily scalarized, use ReverseDiffVJP(true).
		\item If your calculations are on the CPU and your function is very scalarized in operations but has branches, choose ReverseDiffVJP().
		\item If your calculations are on the CPU or GPU and your function is very vectorized, choose ZygoteVJP().
		\item Else fallback to TrackerVJP() if Zygote does not support the function.
		\item If none of the reverse-mode AD based vjps work on your function, fallback to autojacvec=true (for forward-mode AD via ForwardDiff) or false for numerical Jacobians.
	\end{itemize}
\end{itemize}

\section{Continuous vs Discrete Adjoints}\label{SI:adjointtradeoff}

Previous research has shown that the discrete adjoint approach is more stable than continuous adjoints in some cases \cite{onken2020discretize,gholami2019anode,ABRAHAM2004229,betts2005discretize,LIU20141374,callejo2019discrete} while continuous adjoints have been demonstrated to be more stable in others \cite{collis2002analysis,betts2005discretize} and can reduce spurious oscillations \cite{liu2019non,huntley1979note,sirkes1997finite}. This trade-off between discrete and continuous adjoint approaches has been demonstrated on some equations as a trade-off between stability and computational efficiency \cite{hu2018assessment,kepler2010sensitivity,van2005review,kouhi2016implementation,nadarajah2000comparison,gou2011continuous,gauger2015adjoint,c9ce6c121e1949af80032d8978c519cd,DAESCU20035097}. Care has to be taken as the stability of an adjoint approach can be dependent on the chosen discretization method \cite{schwartz1996runge,ghobadi2009discretize,sei1995note,hager2000runge,sandu2005adjoint}, and our software contribution helps researchers switch between all of these optimization approaches in combination with hundreds of differential equation solver methods with a single line of code change.

\section{Integration with Existing Code}

The open-source differential equation solvers of DifferentialEquations.jl \cite{DifferentialEquations.jl-2017} were developed in a manner such that all steps of the programs have a well-defined pullback when using a Julia-based backwards pass generation system. Our software allows for automatic differentiation to be utilized over differential equation solves without any modification to the user code. This enables the simulation software already written with DifferentialEquations.jl, including large software infrastructures such as the MIT-CalTech CLiMA climate modeling system \cite{schneider2017earth} and the QuantumOptics.jl simulation framework \cite{KRAMER2018109}, to be compatible with all of the techniques mentioned in the rest of the paper. Thus while we detail our results in isolation from these larger simulation frameworks, the UDE methodology can be readily used in full-scale simulation packages which are already built on top of the Julia SciML ecosystem. 

\section{Benchmarks \label{SI:Benchmarks}}

\subsection{ODE Solve Benchmarks}

The three ODE benchmarks utilized the Lorenz equations (LRNZ) weather prediction model from \cite{hairerI} and the standard ODE IVP Testset \cite{Testset,testset2}:
\begin{align}
	\frac{dx}{dt} &= \sigma(y-x) \\
	\frac{dy}{dt} &= x(\rho-z) - y \\
	\frac{dz}{dt} &= xy - \beta z
\end{align}
The 28 ODE benchmarks utilized the Pleiades equation (PLEI) celestial mechanics simulation from \cite{hairerI} and the standard ODE IVP Testset \cite{Testset,testset2}:
\begin{align}
	x_i^{\prime\prime} &= \sum_{j\neq i} m_j (x_j - x_i)/r_{ij}\\
	y_i^{\prime\prime} &= \sum_{j\neq i} m_j (y_j - y_i)/r_{ij}
\end{align}
where
\begin{equation}
	r_{ij} = \left((x_i - x_j)^2 + (y_i - y_j)^2 \right)^{3/2}
\end{equation}
on $t \in [0,3]$ with initial conditions:
\begin{align}
	u(0)&=[3.0,3.0,-1.0,-3.0,2.0,-2.0,2.0,3.0,-3.0,2.0,0,0,\\
	&-4.0,4.0,0,0,0,0,0,1.75,-1.5,0,0,0,-1.25,1,0,0]
\end{align}
written in the form $u = [x_i,y_i,x^\prime_i,y^\prime_i]$.

The rest of the benchmarks were derived from a discretization a two-dimensional reaction diffusion equation, representing systems biology, combustion mechanics, spatial ecology, spatial epidemiology, and more generally physical PDE equations:

\begin{align}
	A_t &= D \Delta A + \alpha_A(x) - \beta_A  A - r_1 A B + r_2 C \\
	B_t &= \alpha_B - \beta_B B - r_1 A B + r_2 C \\
	C_t &= \alpha_C - \beta_C C + r_1 A B - r_2 C
\end{align}
where $\alpha_{A}(x) = 1$ if $x>80$ and 0 otherwise on the domain $x \in [0,100]$, $y \in [0,100]$, and $t \in [0,10]$ with zero-flux boundary conditions. For the purpose of parameter gradient tests, we treated calculated the derivative of the solution with respect to the entires of $\alpha_{A}(x)$. The diffusion constant $D$ was chosen as $100$ and all other parameters were left at $1.0$. In this parameter regime the ODE was non-stiff as indicated by solves with implicit methods not yielding performance advantages. The diffusion operator was discretized using the second order finite difference stencil on an $N \times N$ grid, where $N$ was chosen to be 16, 32, 64, 128, 256, and 512. To ensure fairness, the torchdiffeq functions were compiled using torchscript which we varified improved performance. The code for reproducing the benchmark can be found at:
\begin{verbatim}
	https://gist.github.com/ChrisRackauckas/cc6ac746e2dfd285c28e0584a2bfd320
\end{verbatim}

We omit the the gradient performance benchmarks for this case from the main manuscript since the backsolve adjoint method of torchdiffeq is unstable on all of the partial differential equation examples. We note that similarly when using BacksolveAdjoint with the SciML tools we similarly see a divergence, while other adjoints such as InterpolatingAdjoint do not diverge, reinforcing the point that this is due to lack of stability in the algorithm. In the cases where torchdiffeq does not diverge, we see torchdiffeq 12,000x slower and 1,200x slower on Lorenz and Pleiades respectively. However, we note that this is because at the size of those equations forward sensitivity analysis is more efficient which is not available in torchdiffeq, and thus this overestimates the general performance difference in the derivative calculations.

\subsection{Neural ODE Training Benchmark}

The spiral neural ODE from \cite{chen2018neural} was used as the benchmark for the training of neural ODEs. The data was generated according from the form:
\begin{equation}
	u^\prime = Au^3
\end{equation}
where $A = [-0.1,2.0;-2.0,-0.1]$ on $t \in [0,1.5]$ where data was taken at 30 evenly spaced points. Each of the software packages trained the neural ODE for 500 iterations using ADAM with a learning rate of 0.05. The defaults using the SciML software resulted in a final loss of 4.895287e-02 in 7.4 seconds, the optimized version (choosing BacksolveAdjoint with compiled ReverseDiff vector-jacobian products) resulted in a final loss of 2.761669e-02 in 2.7 seconds, while torchdiffeq achieved a final loss of 0.0596 in 289 seconds. To ensure fairness, the torchdiffeq functions were compiled using torchscript. Code to reproduce the benchmark can be found at:
\begin{verbatim}
	https://gist.github.com/ChrisRackauckas/4a4d526c15cc4170ce37da837bfc32c4
\end{verbatim}

\subsection{SDE Solve Benchmark}

The torchsde benchmarks were created using the geometric Brownian motion example from the torchsde README. The SDE was a 4 independent geometric Brownian motions:
\begin{equation}
	dX_t = \mu X_t dt + \sigma X_t dW_t
\end{equation}
where $\mu = 0.5$ and $\sigma = 1.0$. Both software solved the SDE 100 times using the SRI method \cite{doi:10.1137/09076636X} with fixed time step chosen to give 20 evenly spaced steps for $t \in [0,1]$. The SciML ecosystem solvers solved the equation 100 times in 0.00115 seconds, while torchsde v0.1 took 1.86 seconds. We contacted the author who rewrote the Brownian motion portions into C++ and linked it to torchsde as v0.1.1 and this improved the timing to roughly 5 seconds, resulting in a final performance difference of approximately 1,600x. The code to reproduce the benchmarks and the torchsde author's optimization notes can be found at:
\begin{verbatim}
	https://gist.github.com/ChrisRackauckas/6a03e7b151c86b32d74b41af54d495c6
\end{verbatim}

\section{Sparse Identification of Missing Model Terms via Universal Differential Equations}

The SINDy algorithm \cite{bruntonDiscoveringGoverningEquations2016,manganInferringBiologicalNetworks2016,manganModelSelectionDynamical2017} enables data-driven discovery of governing equations from data. Notice that to use this method, derivative data $\dot{X}$ is required. While in most publications on the subject this \cite{bruntonDiscoveringGoverningEquations2016,manganInferringBiologicalNetworks2016,manganModelSelectionDynamical2017} information is assumed. However, for our studies we assume that only the time series information is available. Here we modify the algorithm to apply to only subsets of the equation in order to perform equation discovery specifically on the trained neural network, and in our modification the $\dot{X}$ term is replaced with $U_\theta(t)$, the output of the universal approximator, and thus is directly computable from any trained UDE.

After training the UDE, choose a set of state variables:

\begin{equation}
	\begin{array}{c}
		\mathbf{X}=\left[\begin{array}{c}
			\mathbf{x}^{T}\left(t_{1}\right) \\
			\mathbf{x}^{T}\left(t_{2}\right) \\
			\vdots \\
			\mathbf{x}^{T}\left(t_{m}\right)
		\end{array}\right]=\left[\begin{array}{cccc}
			x_{1}\left(t_{1}\right) & x_{2}\left(t_{1}\right) & \cdots & x_{n}\left(t_{1}\right) \\
			x_{1}\left(t_{2}\right) & x_{2}\left(t_{2}\right) & \cdots & x_{n}\left(t_{2}\right) \\
			\vdots & \vdots & \ddots & \vdots \\
			x_{1}\left(t_{m}\right) & x_{2}\left(t_{m}\right) & \cdots & x_{n}\left(t_{m}\right)
		\end{array}\right] \\
	\end{array}
\end{equation}
and compute a the action of the universal approximator on the chosen states:
\begin{equation}
	\begin{array}{c}
		\dot{\mathbf{X}}=\left[\begin{array}{c}
			\mathbf{x}^{T}\left(t_{1}\right) \\
			\dot{\mathbf{x}}^{T}\left(t_{2}\right) \\
			\vdots \\
			\mathbf{x}^{T}\left(t_{m}\right)
		\end{array}\right]=\left[\begin{array}{cccc}
			\dot{x}_{1}\left(t_{1}\right) & \dot{x}_{2}\left(t_{1}\right) & \cdots & \dot{x}_{n}\left(t_{1}\right) \\
			\dot{x}_{1}\left(t_{2}\right) & \dot{x}_{2}\left(t_{2}\right) & \cdots & \dot{x}_{n}\left(t_{2}\right) \\
			\vdots & \vdots & \ddots & \vdots \\
			\dot{x}_{1}\left(t_{m}\right) & \dot{x}_{2}\left(t_{m}\right) & \cdots & \dot{x}_{n}\left(t_{m}\right)
		\end{array}\right]
	\end{array}
\end{equation}
Then evaluate the observations in a basis $\Theta(X)$. For example:
\begin{equation}
	\Theta(\mathbf{X})=\left[\begin{array}{llllllll}
		1 & \mathbf{X} & \mathbf{X}^{P_{2}} & \mathbf{X}^{P_{3}} & \cdots & \sin (\mathbf{X}) & \cos (\mathbf{X}) & \cdots
	\end{array}\right]
\end{equation}
where $X^{P_i}$ stands for all $P_i$th order polynomial terms such as 
\begin{equation}
	\mathbf{X}^{P_{2}}=\left[\begin{array}{cccccc}
		x_{1}^{2}\left(t_{1}\right) & x_{1}\left(t_{1}\right) x_{2}\left(t_{1}\right) & \cdots & x_{2}^{2}\left(t_{1}\right) & \cdots & x_{n}^{2}\left(t_{1}\right) \\
		x_{1}^{2}\left(t_{2}\right) & x_{1}\left(t_{2}\right) x_{2}\left(t_{2}\right) & \cdots & x_{2}^{2}\left(t_{2}\right) & \cdots & x_{n}^{2}\left(t_{2}\right) \\
		\vdots & \vdots & \ddots & \vdots & \ddots & \vdots \\
		x_{1}^{2}\left(t_{m}\right) & x_{1}\left(t_{m}\right) x_{2}\left(t_{m}\right) & \cdots & x_{2}^{2}\left(t_{m}\right) & \cdots & x_{n}^{2}\left(t_{m}\right)
	\end{array}\right]
\end{equation}

Using these matrices, find this sparse basis $\mathbf{\Xi}$ over a given candidate library $\mathbf{\Theta}$ by solving the sparse regression problem $\dot{X} =\mathbf{\Theta}\mathbf{\Xi}$ with $L_1$ regularization, i.e. minimizing the objective function $\left\Vert \mathbf{\dot{X}} - \mathbf{\Theta}\mathbf{\Xi} \right\Vert_2 + \lambda \left\Vert \mathbf{\Xi}\right\Vert_1$. This method and other variants of SINDy applied to UDEs, along with specialized optimizers for the LASSO $L_1$ optimization problem, have been implemented by the authors and collaborators as the DataDrivenDiffEq.jl library on top of the ModelingToolkit.jl and Symbolics.jl computer algebra systems.

\subsection{Application for the Reconstruction of Unknown Dynamics \label{SI:LV}}
\subsubsection{Application to the Partial Reconstruction of the Lotka-Volterra \label{SI:LV_Partial}}

On the Lotka-Volterra equations, we trained a UDE model in two different scenarios,  starting from $x_0= 0.44249296,~y_0= 4.6280594$  with parameters are chosen to be  $\alpha = 1.3,~ \beta =  0.9,~ \gamma = 0.8,~ \delta = 1.8$. The neural network consists of an input layer, two hidden layers with 5 neurons and a linear output layer modeling the polynomial interaction terms. The input and hidden layer have gaussian radial basis activation functions. We trained for 200 iterations with ADAM with a learning rate $\gamma = 10^{-1}$. We then switched to BFGS with an initial stepnorm of $\gamma = 10^{-2}$ setting the maximum iterations to 10000. Typically the training converged after 400-600 iterations in total. 

\textit{Scenario 1)} consists of a trajectory with 31 points measured in $x(t), y(t)$ with a constant step size $\Delta t = 0.1$. We assumed perfect knowledge about the linear terms of the equations and their corresponding parameters $\alpha,~\delta$. The trajectory has been perturbed with additive noise drawn from a normal distribution scaled by $5\%$ of its mean. The loss was chosen was the L2 loss $\mathcal{L} = \sum_i (u_\theta(t_i) - d_i)^2$.

\begin{figure}
	\centering
	\includegraphics[width = \textwidth]{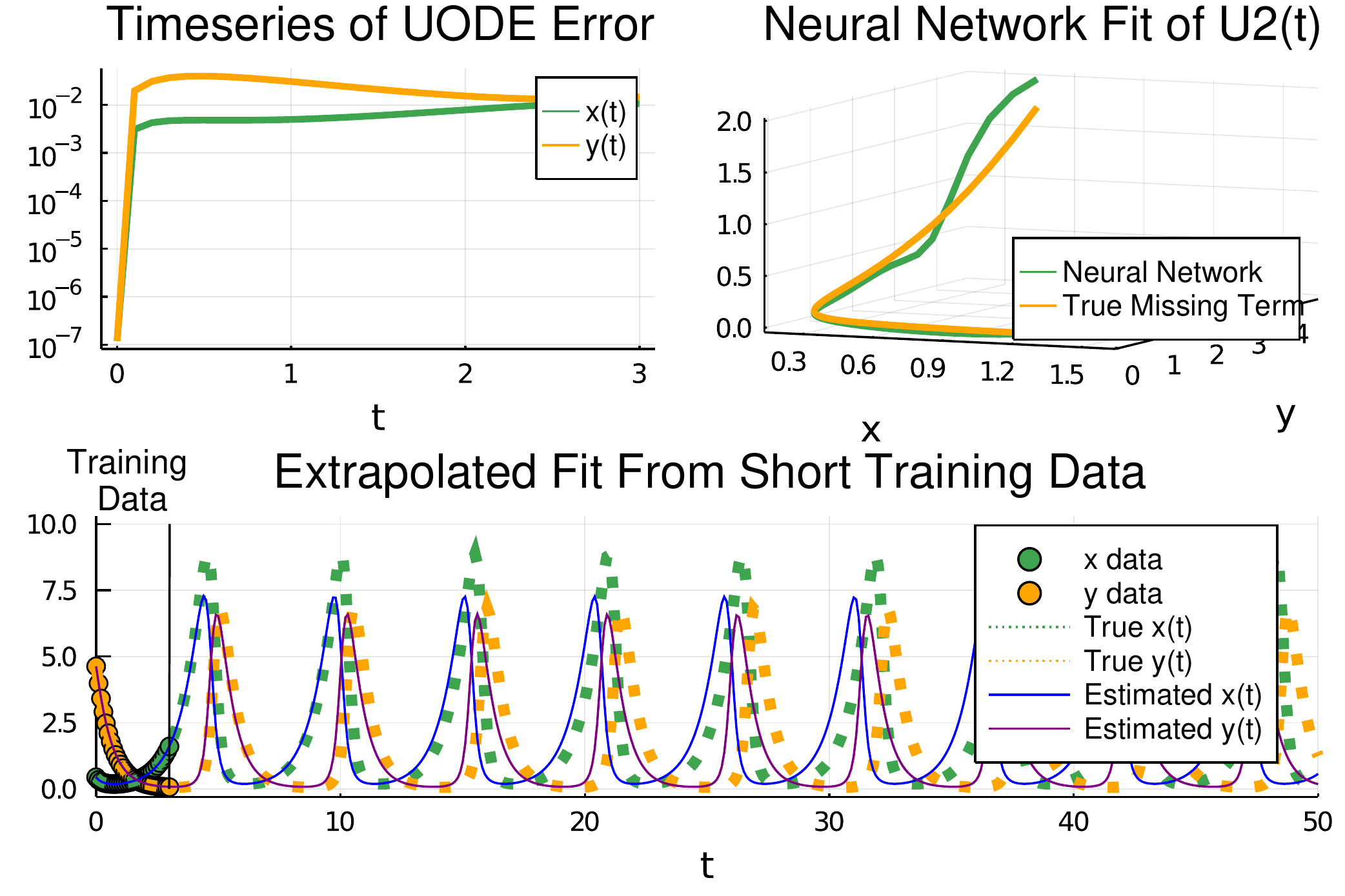}
	\caption{\textit{Scenario 1)} of the Lotka Volterra Recovery}
	\label{SIfig:LV_Scenario1}
\end{figure}

\textit{Scenario 2)} consists of a trajectory with 61 points measured in  $x(t)$ with a constant step size $\Delta t = 0.1$ and 6 points measured in $y(t)$ with a constant step size $\Delta t = 1.2$. The trajectory has been perturbed with additive noise drawn from a normal distribution scaled by $1\%$ of its mean.We assumed perfect knowledge about the linear terms of the first differential equation and their corresponding parameters $\alpha$. In addition to the UDE a linear decay rate with unknown parameter was added in the second differential equation governing the predator dynamics. The parameter was included in the training process. The data was divided into $j=5$ different datasets $d$ containing $i=13$ measurements in $x,~t$ and 2 measurements in $y$ for the initial and boundary condition. The loss was chosen was similar to shooting like techniques plus a regularization term $\mathcal{L} = \sum_j (\sum_i(u_{x,\theta(t_i,j)} - d_{x,i,j})^2 + \|u_{y,\theta(t_13,j)} - d_{y,2,j}\|) + \lambda \|\theta\|^2$.

\begin{figure}
	\centering
	\includegraphics[width = \textwidth]{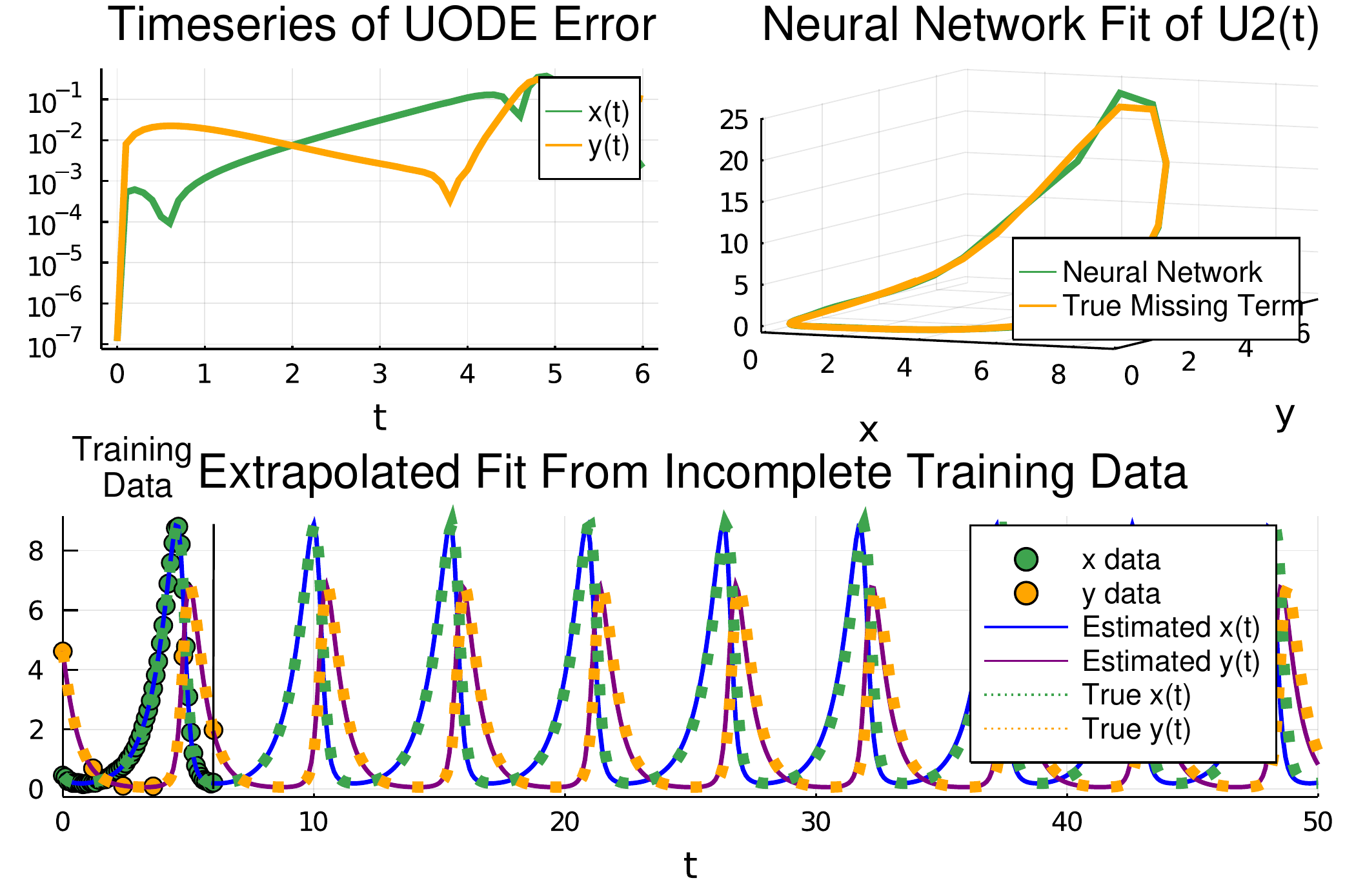}
	\caption{\textit{Scenario 2)} of the Lotka Volterra Recovery}
	\label{SIfig:LV_Scenario2}
\end{figure}

From the trained neural network, data was sampled over the original trajectory and fitted using the SR3 method with varying threshold with $\lambda = exp10.(-7:0.1:3)$. A pareto optimal solution was selected via the L1 norm of the coefficients and the L2 norm of the corresponding error between the differential data and estimate. The knowledge-enhanced neural network returned zero for all terms except non-zeros on the quadratic terms with nearly the correct coefficients that were then fixed using an additional sparse regression over the quadratic terms. The resulting parameters extracted are $\beta \approx 0.9239$ and $ \gamma \approx 0.8145$. Performing a sparse identification on the ideal, full solution and numerical derivatives computed via an interpolating spline resulted in a failure. After determining the results of the symbolic regression, the learned ODE model was then trained to refit the parameters before extrapolating.
In comparison, interpolations of the sparse predator measurements have been performed using linear, quadratic and cubic spline and a polynomial interpolation of order 5. 


Using SINDy with same library of candidate functions over the interpolated data and their numerical derivatives leads to the the results listed in \ref{SItab:LV_Scenario2_ITP}. Given the sparsity of the data, quality of the resulting fit and its derivative, none of the attempts were able to recover the true equation describing the predator dynamics.

\begin{table}[h]
	\centering
	\begin{tabular}{|l|l|}
		\hline
		Interpolation & Recovered Equation  \\ \hline
		Linear & $\dot{y} = 2.19 \cdot x^{2} - 0.64 \cdot x^{3} + 0.05 \cdot x^{4} - 0.01 \cdot y^{4} $\\
		Quadratic Spline & $\dot{y} = - 0.77 \cdot y^{3} + 0.23 \cdot y^{4} - 0.02 \cdot y^{5}$ \\
		Cubic Spline & $\dot{y} =0.69 \cdot x^{3} - 0.14 \cdot x^{4} + 0.03 \cdot x^{4} \cdot y - 0.02 \cdot x^{3} \cdot y^{2}$ \\
		Polynomial & $\dot{y} = 0.08 \cdot x^{3} - 0.03 \cdot x^{4} + 9.62 \cdot y^{3} - 2.09 \cdot y^{4} + 0.06 \cdot y^{5}$ \\
		\hline
	\end{tabular}
	\caption{Results of using SINDy on interpolation data of the predator measurements}
	\label{SItab:LV_Scenario2_ITP}
\end{table}

\begin{figure}
	\centering
	\includegraphics[width=\textwidth]{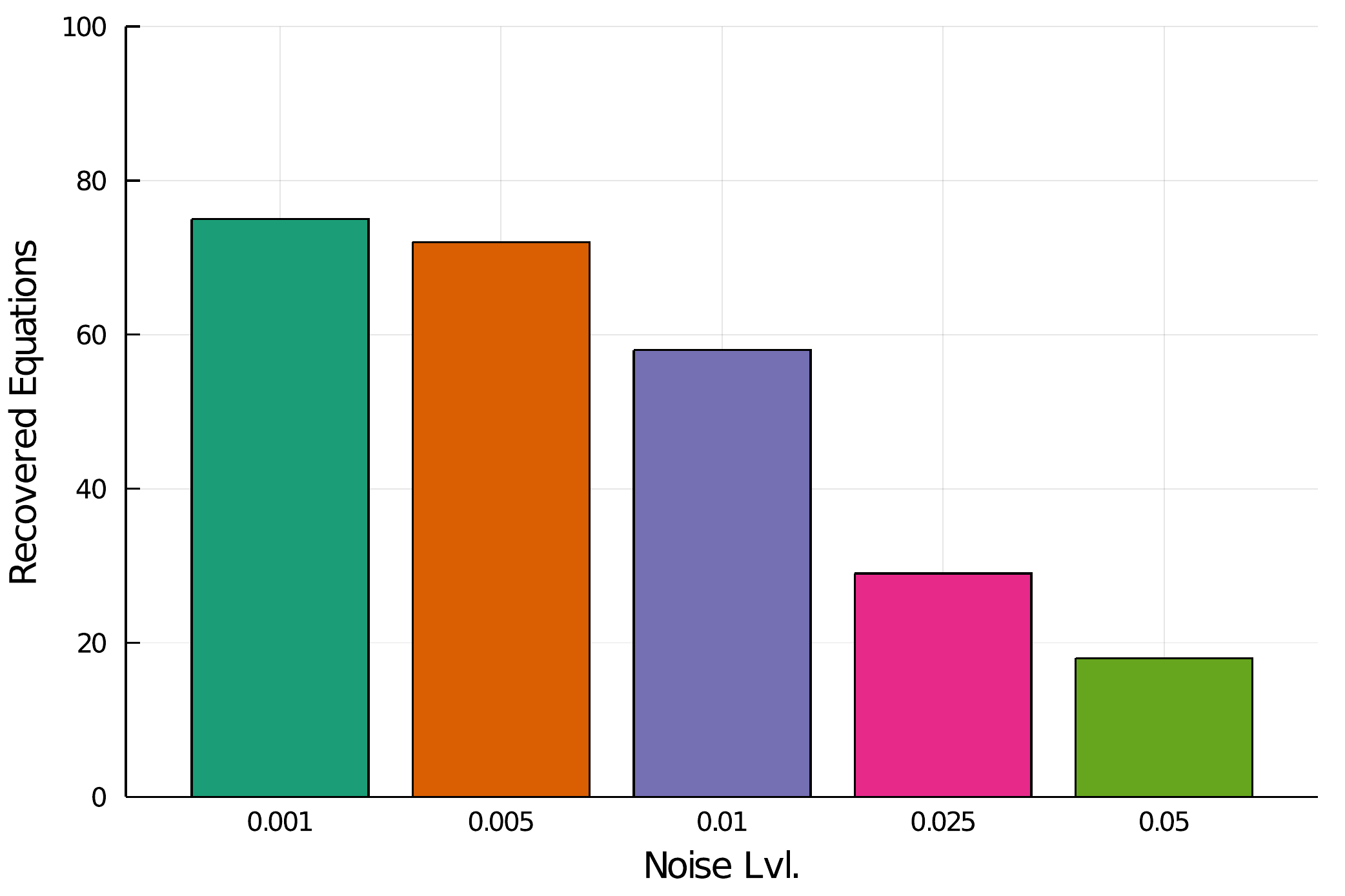}
	\caption{Successful recoveries of the partial reconstruction of the Lotka Volterra equations with varying noise level with the sparse data.}
	\label{SIfig:LV_Rec}
\end{figure}
\begin{figure}
	\centering
	\includegraphics[width=\linewidth]{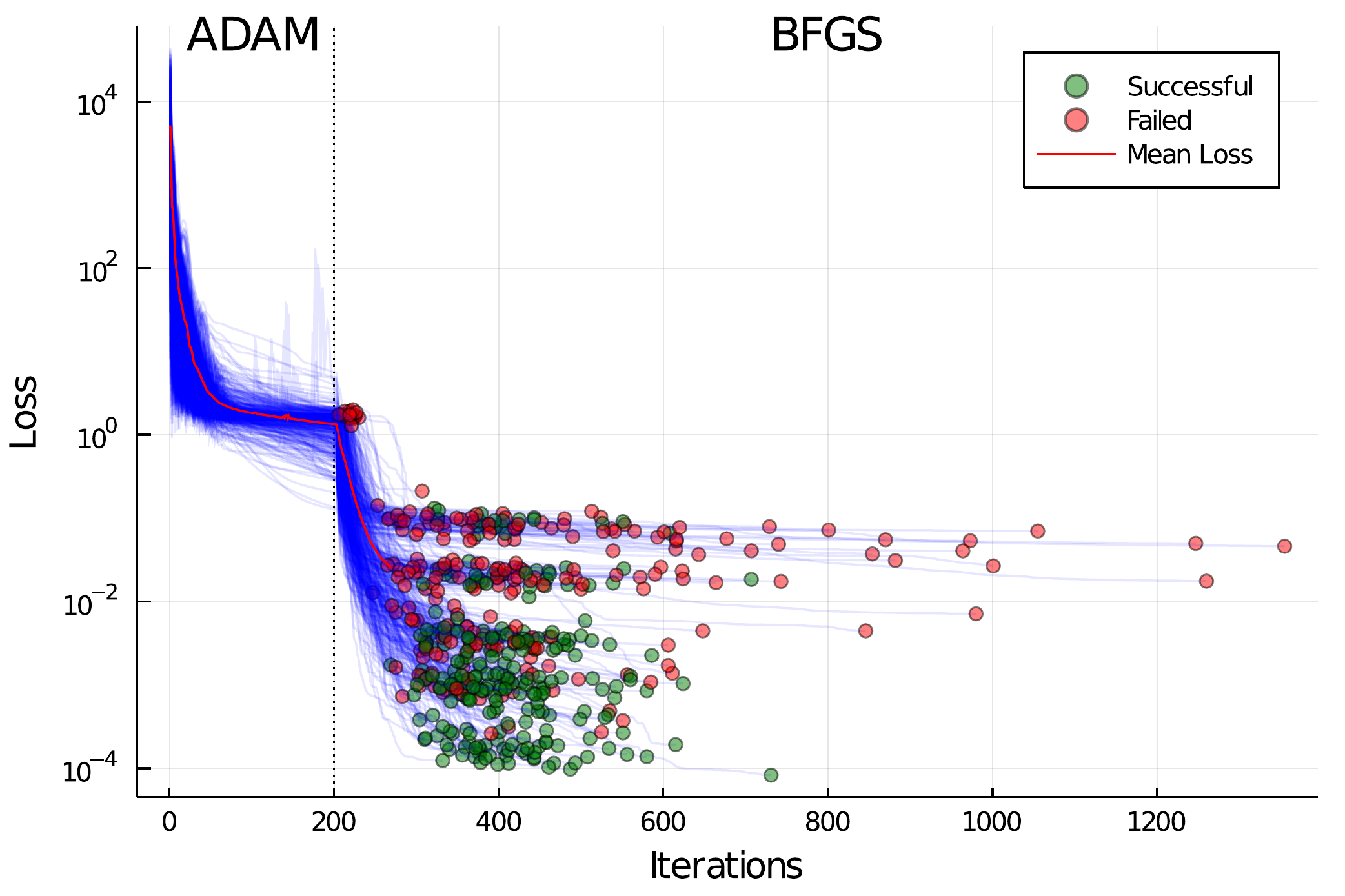}
	\caption{Loss trajectories for 498 runs of the partial reconstruction of the Lotka Volterra equations with varying noise level.}
	\label{SIfig:LV_Loss}
\end{figure}

The UDE training procedure was evaluated 500 times in total for \textit{Scenario 1)} given the initial trajectory with varying noise levels $0.1, ~ 0.5, ~1.0,~ 2.5,~5\%$ in terms of the mean of the trajectory (100 repetitions each). The results of the successful recovery of the missing equations are shown in  Fig. \ref{SIfig:LV_Rec}, the training loss is shown in Fig. \ref{SIfig:LV_Loss}. Overall, the recovery rate is $(50.4 \pm 25.7) \%$ for all 498 error-free runs . Two runs failed due to numerical instabilities of the combination of trained parameters and numerical integrator. Given that the study has conducted in a brute-force manor, e.g. no prior data cleaning, no hyperparameter optimization, no (automatic) pre-selection of the networks architecture or candidate selection has been performed, the sensitivity of the networks initial parameters is also captured in this study.
However, since the scope of this work is to highlight the usefulness of UDEs as means to extract specific information from time series we limited ourselves to this naive approach. Further research can highlight additional methods to improve noise robustness.

\begin{figure}[!h]
	\centering
	\includegraphics[width=\linewidth]{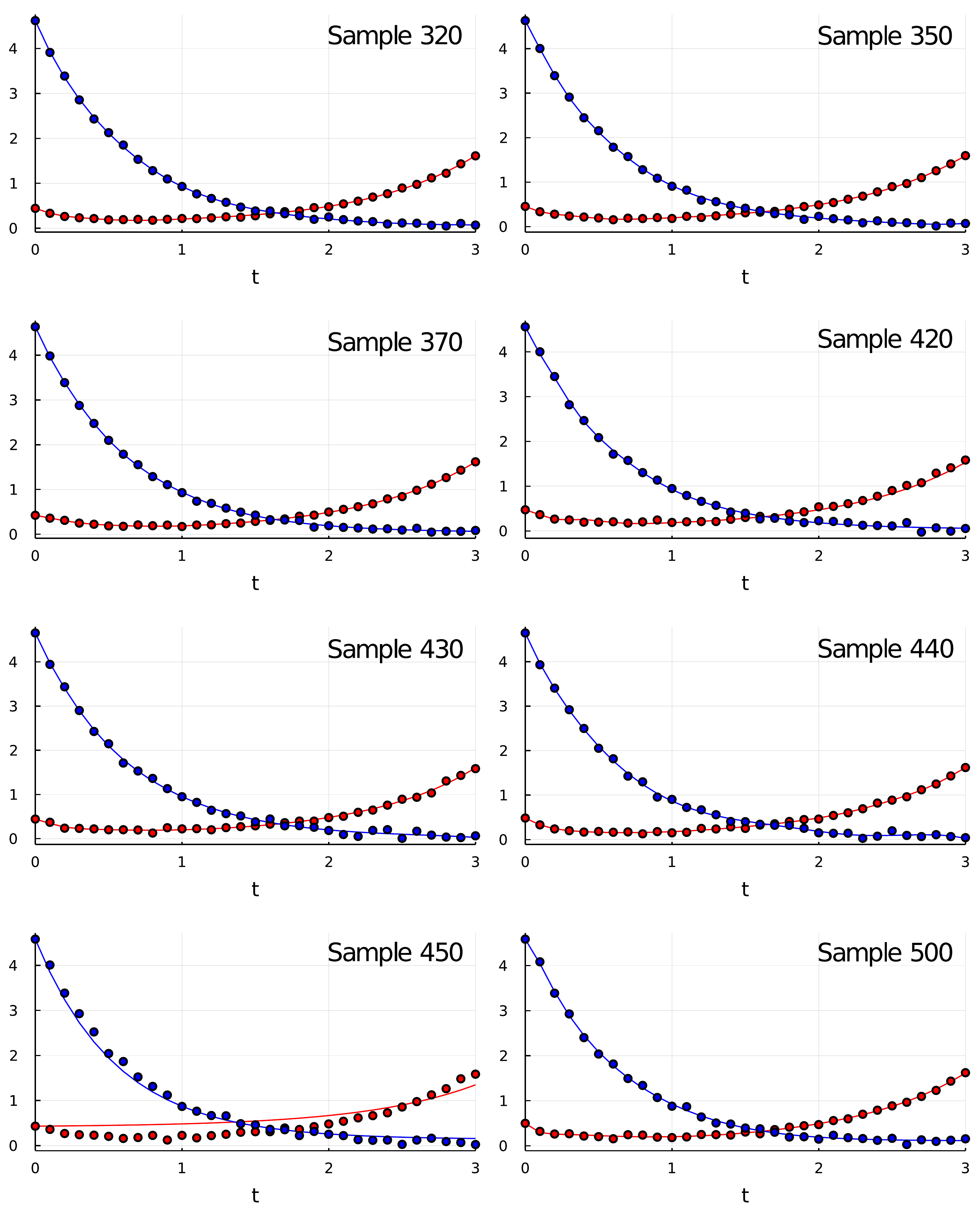}
	\caption{Examples of failed recovery of the Lotka Volterra Equations.}
	\label{SIfig:LV_Failed}
\end{figure}

In Fig. \ref{SIfig:LV_Failed} some examples for failed recovery can be seen. It is worth noting that the mere visual quality of the fit points in most cases, except for Sample 450, would indicate a success. In some cases, e.g. Sample 500, some discontinuities can be seen in the solution, possibly due to overfitting. However, adding a regularization penalty to the overall loss function has shown little effect and was hence neglected. 

\subsubsection{Application to the Reconstruction of the Lotka-Volterra Using the Hudson Bay Dataset \label{SI:LV_Hudson}}
Additionally, a full recovery based on a dataset of the \textit{Hudson Bay} Data of Hares and Lynx between 1900 and 1920, taken from \cite{HudsonBay} and originally published in \cite{Odum1953} has been performed. The data has been normalized to the interval $(0,1]$ and the assumed model is given as

\begin{equation}
	\begin{aligned}
		\dot{x} &= \theta_1 x + U_1(\theta_{3\cdots}, x,y)\\
		\dot{y} &= -\theta_2 y + U_2(\theta_{3\cdots}, x,y)
	\end{aligned}
\end{equation}

\begin{figure}[H]
	\centering
	\includegraphics[width = \linewidth]{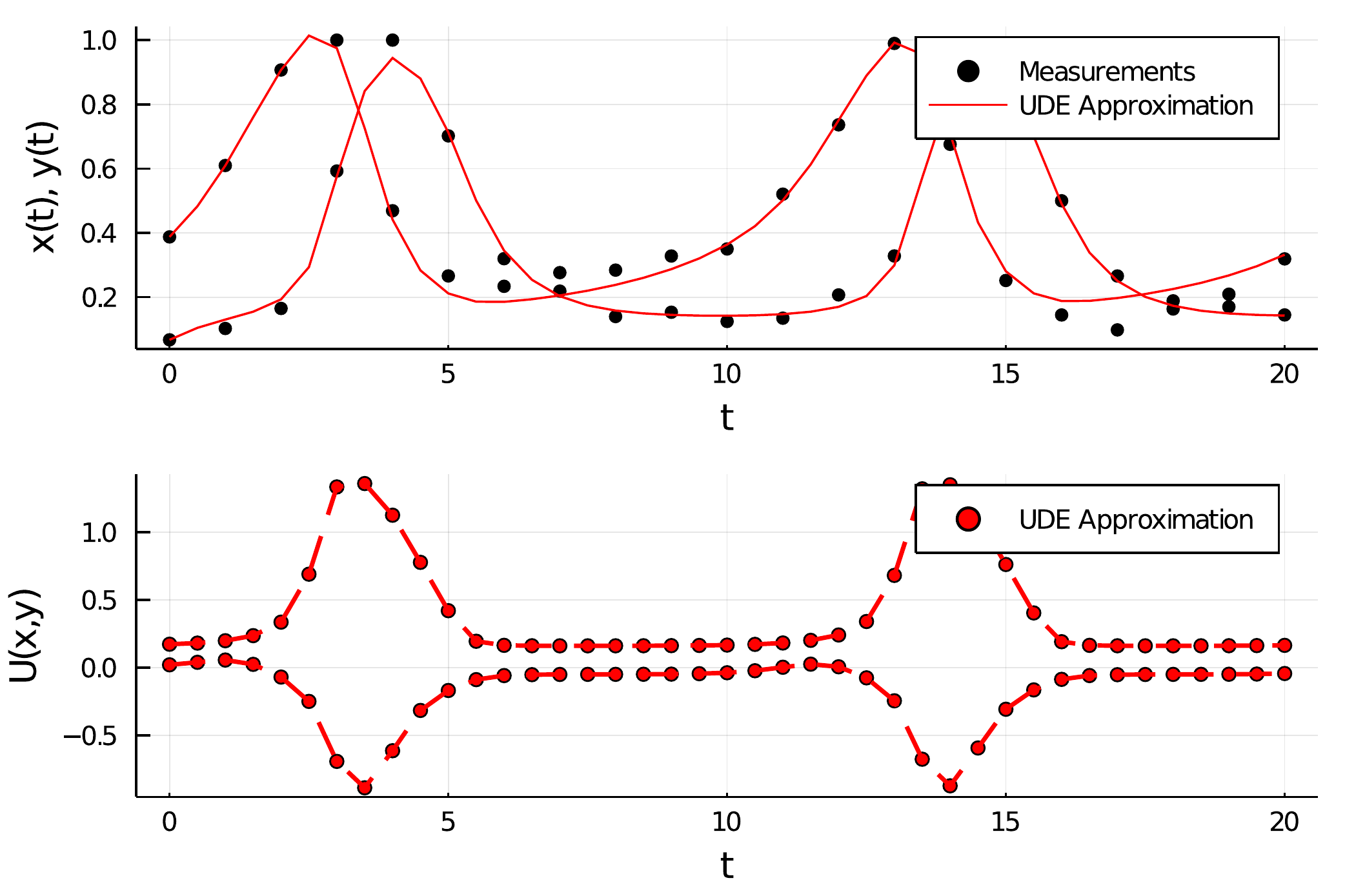}
	\caption{Application of UDE training to Hudson Bay data.}
	\label{SIfig:LV_Hudson_1}
\end{figure}

Incorporating a linear birth rate of the prey and a linear decay rate of the predator. The neural network consists of an input layer, two hidden layers with 5 neurons and a linear output layer. Again, radial basis activations have been used, except for the last hidden layer which used a tanh activation. As in \textit{Scenario 2)} we started by using a shooting loss with L2 regularization of the parameter, intentionally widthholding data of the predators to smoothly optimize the parameters using ADAM with a learning rate of $0.1$ for 100 iterations and BFGS with an initial stepnorm of $0.01$ until converged ( maximum 200 iterations). After this initial fit, we switched onto a an L2-Norm loss with regularization and trained until convergence using BFGS. The results are shown in Figure \ref{SIfig:LV_Hudson_1}.

\begin{figure}[H]
	\centering
	\includegraphics[width = 0.9\linewidth]{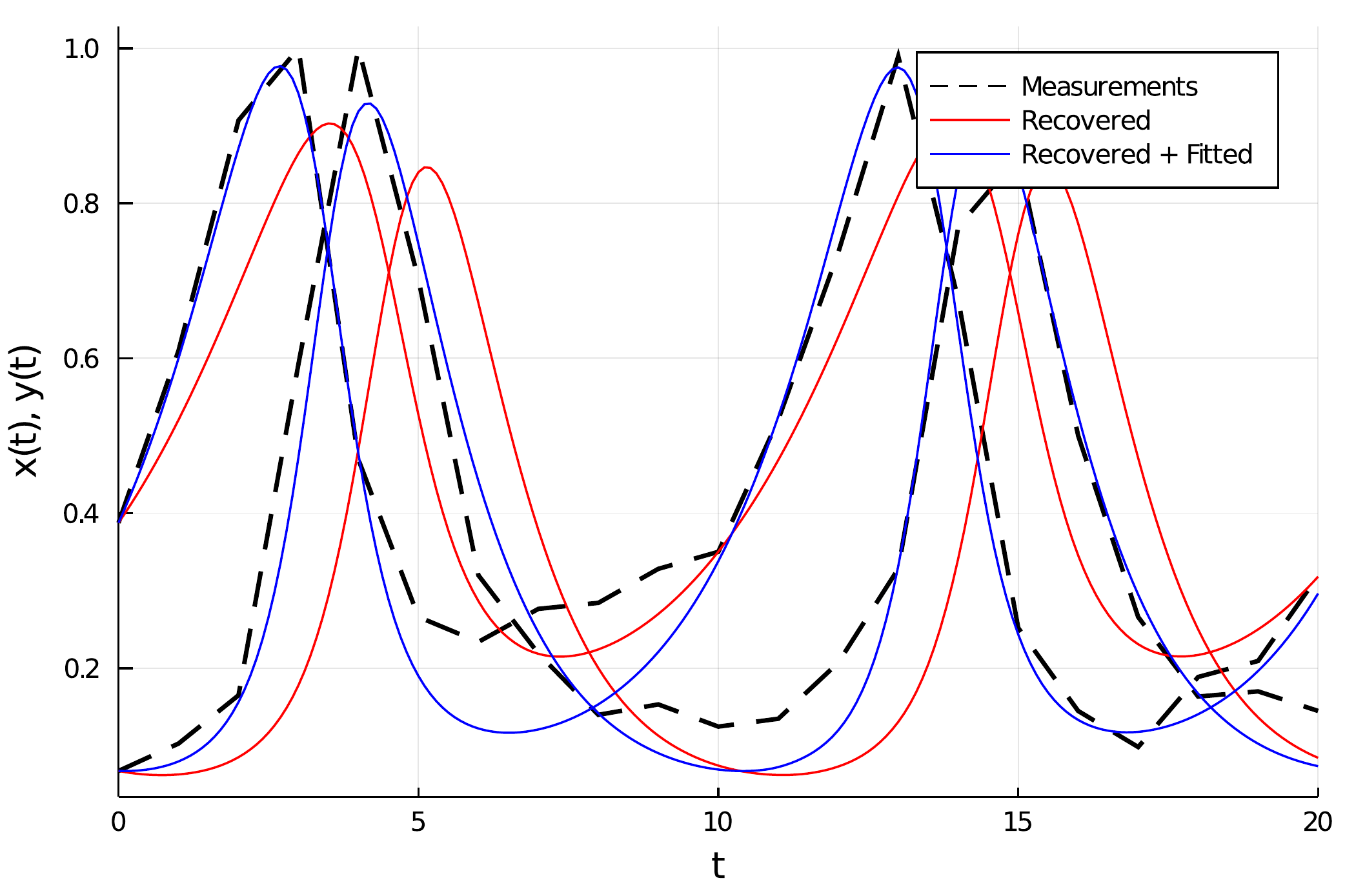}
	\caption{Hudson Bay Dataset (black), the recovered equations before (red) and after (blue) parameter fitting}
	\label{SIfig:LV_Hudson_2}
\end{figure}

\begin{figure}[H]
	\centering
	\includegraphics[width = 0.9\linewidth]{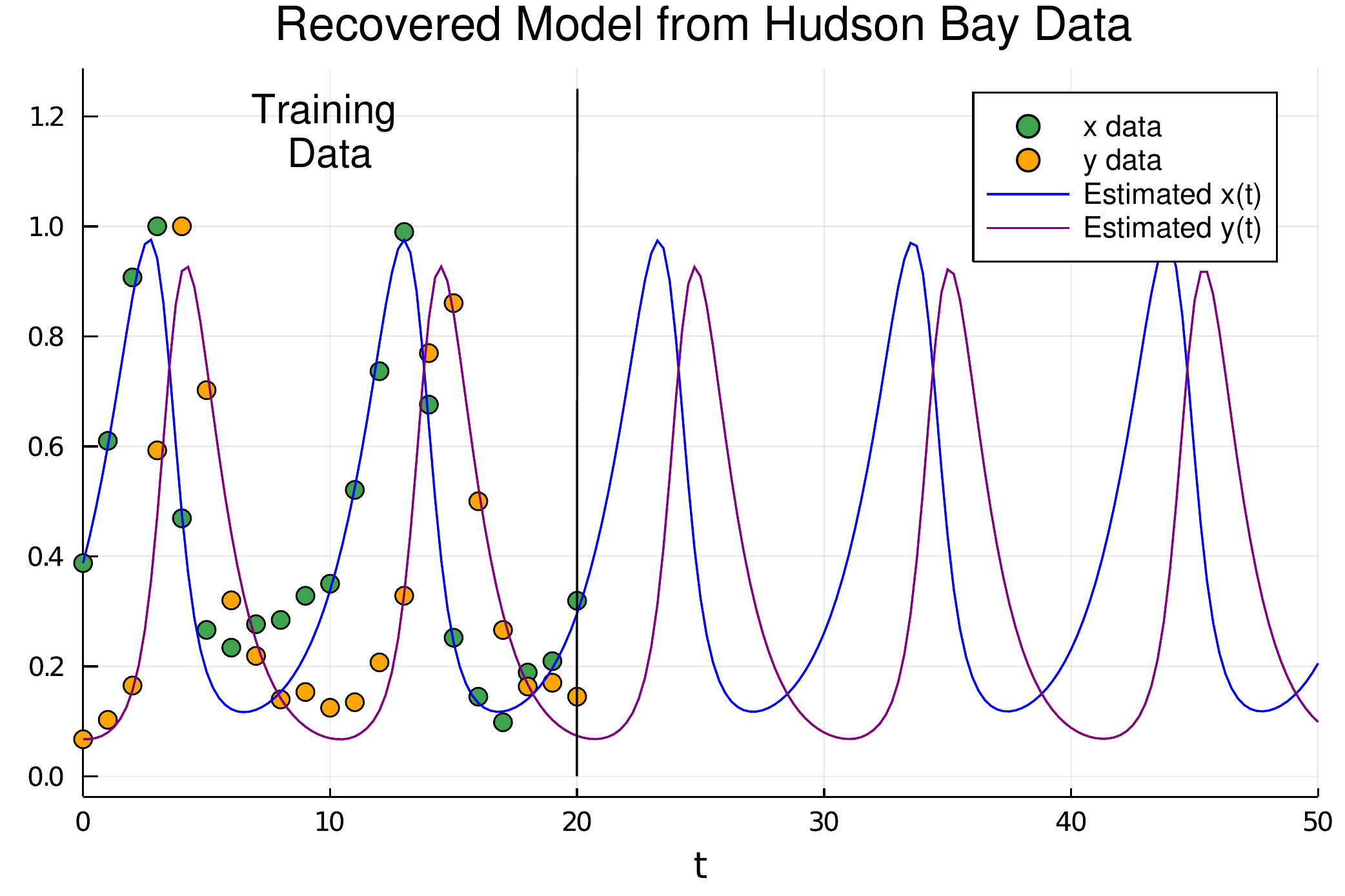}
	\caption{Resulting Model recovered from the Hudson Bay data and its extrapolation}
	\label{SIfig:LV_Hudson_3}
\end{figure}

Afterwards we performed a sparse regression with a candidate library of multivariate polynomials up to degree 5 and sinusoidal signals in the states on the recovered signal of the UDE using sequentially thresholded regression. The UDE has been subsampled to an interval of 0.5 years, efficcently augmenting the limited data. The mixed terms were recovered successful. The resulting symbolic model has been post-fitted with an L2-Norm loss to decrease its resulting error, as can be seen in Figure \ref{SIfig:LV_Hudson_2}. A long term estimation can seen in Figure \ref{SIfig:LV_Hudson_3}, using the parameters $\alpha = 0.557$, $\delta = 0.826$, $\beta = -1.70$ and $\gamma = 2.04$.

\subsubsection{Application to the Reconstruction of the Fisher-KPP Equations \label{SI:Fisher_Reconstruction}}

To generate training data for the 1D Fisher-KPP equation \ref{Eqn:Fisher-KPP}, we take the growth rate and the diffusion coefficient to be $r=1$ and $D = 0.01$ respectively. The equation is numerically solved in the domain $x \in [0, 1]$ and $t \in [0, T]$ using a 2nd order central difference scheme for the spatial derivatives and the time-integration is done using the Tsitouras 5/4 Runge-Kutta method.  We implement periodic boundary condition $\rho(x=0, t) = \rho(x=1, t)$ and initial condition $\rho(x, t=0) = \rho_0(x)$ is taken to be a localized function given by 
\begin{equation}
	\rho_0(x) = \frac{1}{2}\left( \tanh\left(\frac{x - (0.5 - \Delta/2)}{\Delta/10}\right) - \tanh\left(\frac{x - (0.5 + \Delta/2)}{\Delta/10}\right)\right),
	\label{Eqn:Fisher-IC}
\end{equation}
with $\Delta = 0.2$ which represents the width of the region where $\rho \simeq 1$. The data are saved at evenly spaced points with $\Delta x = 0.04$ and $\Delta t = 0.5$.

In the UPDE \ref{Eqn:UPDE-Fisher}, the growth neural network $\textrm{NN}_{\theta}(\rho)$ has 4 densely connected layers with 10, 20, 20 and 10 neurons each and $\tanh$ activation functions. The diffusion operator is represented by a CNN that operates on an input vector of arbitrary size. It has 1 hidden layer with a $3 \times 1$ filter $[w_1, w_2, w_3]$ without any bias. To implement periodic boundary conditions for the UPDE at each time step, the vector of values at different spatial locations $[\rho_1, \rho_2, \dots , \rho_{N_x}]$ is padded with $\rho_{N_x}$ to the left and $\rho_1$ to the right. This also ensures that the output of the CNN is of the same size as the input. The weights of both the neural networks and the diffusion coefficient are simultaneously trained to minimize the loss function  
\begin{equation}
	\mathcal{L} = \sum_{i} (\rho (x_i, t_i) - \rho_{\textrm{data}}(x_i, t_i))^2 + \lambda |w_1 + w_2 + w_3|,
\end{equation}
where $\lambda$ is taken to be $10^2$ (note that one could also structurally enforce $w_3 = -(w_1 + w_2))$. The second term in the loss function enforces that the differential operator that is learned is conservative---that is, the weights sum to zero. The training is done using the ADAM optimizer with learning rate $10^{-3}$.

\begin{figure}[H]
	\centering
	\includegraphics[width = 0.9\linewidth]{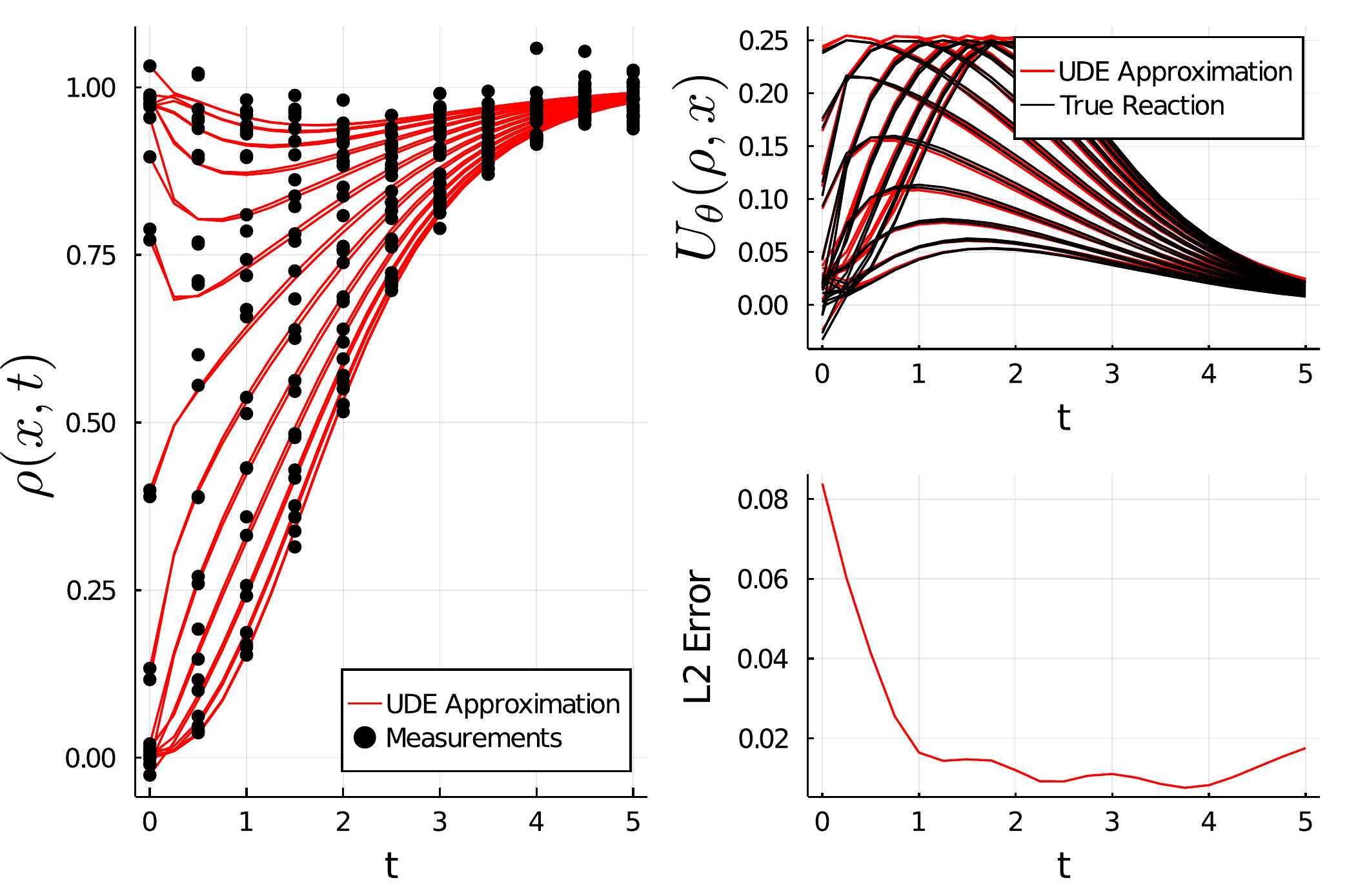}
	\caption{Recovered reaction term (red) of the one-dimensional Fisher-KPP with periodic boundary conditions with noisy measurements (black) and the overall L2-Norm Error}
	\label{fig:fisher-kpp_reconstruction}
\end{figure}

Similar to the Lotka-Volterra experiments, symbolic regression over monomials up to degree 10 has been performed to recover the nonlinear term of the Fisher-KPP Equations $U_\theta$ from Section \ref{SI:KPPUA} in \textit{Scenario 3)}. The system has been simulated with $r = 1$ for $t = [0, 5] s$ with a time resolution of $\Delta t = 0.5 s$ and 25 measurements in $x$. Normally distributed noise with $2.5\%$ of the mean has been added. Figure \ref{fig:fisher-kpp_reconstruction} shows the recovered dynamics nonlinear term via the UDE approach, using a network similar to \textit{Scenario 1)}. We trained the network using a loss function similar to Section \ref{SI:KPPUA} with a regularization of $1$ using ADAM with a learning rate of $0.1$ for 200 iterations and then switching to BFGS with an initial stepnorm of $0.01$. The recovered equation is $U_\theta(\rho) = 1.0\rho - 1.0\rho^2$. The symbolic regression has been performed using sequentially thresholded least squares over thresholds $\lambda = [10^{-3}, 10^2]$ logarithmic evenly distributed with a step size of $0.01$.

\subsubsection{Analysis of Alternative Function Approximators in Fisher-KPP \label{SI:KPPUA}}

Additionally, we analyzed the performance of alternative universal approximators for this spatiotemporal PDE discovery task. To do this, we analyzed both the parameter efficiency and the computational efficiency. For the parameter efficiency, we attempted to quantify the minimum numbers of parameters that could robustly identify the terms of the PDEs. It is predicted by the lottery ticket hypothesis \cite{frankle2018lottery} that low parameter neural networks can generally produce accurate fits, but have a low probability of being discovered. Thus we defined robustness as the ability to perform 5 optimizations with random initializations mixed with stochastic optimizers and recover a fitted solution to a loss of 0.01. All optimizations used the same optimization process which was 400 iterations of ADAM to find a local minima and then BFGS to complete the optimization (which automatically exited when the solution achieves a local maximum according to the default detection criteria of the Optim.jl library \cite{Optim.jl-2018}). 

For the neural network we choose 1 hidden layer with a $\tanh$ activation function and scaled the size of the hidden layer from 2-4, corresponding to 4, 7, and 15 parameters. At 4 parameters, all 5 optimization runs failed to produce a loss of 0.01, while at 7 and 15 all optimizations passed. To test against a non-neural network function approximator we chose to use the Fourier basis. Tests with 3, 5, 7, and 15 parameters all give fits to a loss of 0.01 on all 5 optimization runs, demonstrating that the smaller parameter Fourier basis model was more robust than the smallest parameter neural network model. The least parameter neural network, the 7 parameter version, took approximately 2,500 seconds with a standard deviation of approximately 1,000 seconds, a minimum time of approximately 1,300 seconds and a maximum time of approximately 3,300 seconds. The Fourier basis with 7 parameters took approximately 250 seconds to train, with a standard deviation of approximately 5 seconds, a minimum time of approximately 242 seconds and a maximum time of approximately 255 seconds. This demonstrates that the training time with the Fourier basis was on average an order of magnitude faster than the neural network when comparing the same parameter size between the models.

All of the output losses and times for the experiments are stored as comments in the code for the experiments Fisher\-KPP\-CNN\-Small.jl and Fisher\-KPP\-CNN\-Fourier.jl in the example repository. The implementations of the classical basis functions can be found in the DiffEqFlux.jl repository \footnote{\url{https://diffeqflux.sciml.ai/dev/layers/BasisLayers/}} with an associated tutorial on classical basis functions and TensorLayer \footnote{\url{https://diffeqflux.sciml.ai/dev/examples/tensor_layer/}}.

\subsection{Discovery of Robertson's Equations with Prior Conservation Laws \label{SI:Robertson}}

On Robertson's equations, we trained a UDAE model against a trajectory of 10 points on the timespan $t \in [0.0,1.0]$ starting from $y_1 = 1.0$, $y_2 = 0.0$, and $y_3 = 0.0$. The parameters of the generating equation were $k_1 = 0.04$, $k_2 = 3e7$, and $k_3 = 1e4$. The universal approximator was a neural network with one hidden layers of size 64. The equation was trained using the BFGS optimizer to a loss of $9e-6$.

\section{Adaptive Solving for the 100 Dimensional Hamilton-Jacobi-Bellman Equation \label{SI:HJB}}

\subsection{Forward-Backwards SDE Formulation}

Consider the class of semilinear parabolic PDEs, in finite time $t\in[0, T]$ and $d$-dimensional space $x\in\mathbb R^d$, that have the form:

\begin{equation}
	\begin{aligned}
		\frac{\partial u}{\partial t}(t,x) 	&+\frac{1}{2}\text{tr}\left(\sigma\sigma^{T}(t,x)\left(\text{Hess}_{x}u\right)(t,x)\right)\\
		&+\nabla u(t,x)\cdot\mu(t,x) \\
		&+f\left(t,x,u(t,x),\sigma^{T}(t,x)\nabla u(t,x)\right)=0,
	\end{aligned}
\end{equation}
with a terminal condition $u(T,x)=g(x)$. In this equation, $\text{tr}$ is the trace of a matrix, $\sigma^T$ is the transpose of $\sigma$, $\nabla u$ is the gradient of $u$, and $\text{Hess}_x u$ is the Hessian of $u$ with respect to $x$. Furthermore, $\mu$ is a vector-valued function, $\sigma$ is a $d \times d$ matrix-valued function and $f$ is a nonlinear function. We assume that $\mu$, $\sigma$, and $f$ are known. We wish to find the solution at initial time, $t=0$, at some starting point, $x = \zeta$.

Let $W_{t}$ be a Brownian motion and take $X_t$ to be the solution to the stochastic differential equation
\begin{equation}\label{Xsde}
	dX_t = \mu(t,X_t) dt + \sigma (t,X_t) dW_t
\end{equation}
with a terminal condition $u(T,x)=g(x)$. With initial condition $X(0)=\zeta$ has shown that the solution to \ref{pdeform} satisfies the following forward-backward SDE (FBSDE) \cite{zhang2017backward}:

\begin{align}
	\label{ubsde}
	u(t, &X_t) - u(0,\zeta) = \nonumber\\
	& -\int_0^t f(s,X_s,u(s,X_s),\sigma^T(s,X_s)\nabla u(s,X_s)) ds \nonumber\\
	& + \int_0^t \left[\nabla u(s,X_s) \right]^T \sigma (s,X_s) dW_s,
\end{align}
with terminating condition $g(X_T) = u(X_T,W_T)$. Notice that we can combine \ref{Xsde} and \ref{ubsde} into a system of $d+1$ SDEs:

\begin{equation}
	\begin{aligned}\label{eq:composite_sde}
		dX_t =& \mu(t,X_t) dt + \sigma (t,X_t) dW_t,\\
		dU_t =& f(t,X_t,U_t,\sigma^T (t,X_t) \nabla u(t,X_t)) dt\\
		&+ \left[\sigma^T (t,X_t) \nabla u(t,X_t)\right]^T dW_t,
	\end{aligned}
\end{equation}
where $U_t = u(t,X_t)$. Since $X_0$, $\mu$, $\sigma$, and $f$ are known from the choice of model, the remaining unknown portions are the functional $\sigma^T (t,X_t) \nabla u(t,X_t)$ and initial condition $U(0) = u(0,\zeta)$, the latter being the point estimate solution to the PDE.

To solve this problem, we approximate both unknown quantities by universal approximators:

\begin{equation}
	\begin{aligned}\label{eq:nns}
		\sigma^T (t,X_t) \nabla u(t,X_t) &\approx U^1_{\theta_1}(t,X_t), \\
		u(0,X_0) &\approx U^2_{\theta_2}(X_0),
	\end{aligned}
\end{equation}
Therefore we can rewrite \ref{eq:composite_sde} as a stochastic UDE of the form:
\begin{equation}
	\begin{aligned}
		dX_t &= \mu(t,X_t) dt + \sigma (t,X_t) dW_t,\\
		dU_t &= f(t,X_t,U_t,U^1_{\theta_1}(t,X_t)) dt + \left[U^1_{\theta_1}(t,X_t)\right]^T dW_t,
	\end{aligned}
\end{equation}
with initial condition $(X_0,U_0)=(X_0,U^2_{\theta_2}(X_0))$.

To be a solution of the PDE, the approximation must satisfy the terminating condition, and thus we define our loss to be the expected difference between the approximating solution and the required terminating condition:

\begin{equation}
	\label{lossfn}
	l(\theta_1,\theta_2|X_T, U_T) = \mathbb{E} \left[\left\Vert g(X_T) - U_T \right\Vert\right].
\end{equation}

Finding the parameters $(\theta_1,\theta_2)$ which minimize this loss function thus give rise to a BSDE which solves the PDE, and thus $U^2_{\theta_2}(X_0)$ is the solution to the PDE once trained.

\subsection{The LQG Control Problem}

This PDE can be rewritten into the canonical form by setting:

\begin{equation}
	\begin{aligned}
		\mu &= 0, \\
		\sigma &= \overline{\sigma} I, \\
		f &= -\alpha \left \| \sigma^T(s,X_s)\nabla u(s,X_s)) \right \|^{2},
	\end{aligned}
\end{equation}
where $\overline{\sigma} = \sqrt{2}$, T = 1 and $X_0 = (0,. . . , 0) \in R^{100}$. The universal stochastic differential equation was then supplemented with a neural network as the approximator. The initial condition neural network was had 1 hidden layer of size 110, and the $\sigma^T (t,X_t) \nabla u(t,X_t)$ neural network had two layers both of size 110. For the example we chose $\lambda = 1$. This was trained with the LambaEM method of DifferentialEquations.jl \cite{DifferentialEquations.jl-2017} with relative and absolute tolerances set at $1e-4$ using 500 training iterations and using a loss of 100 trajectories per epoch. 

On this problem, for an arbitrary $g$, one can show with It\^o's formula that:

\begin{equation}
	u(t,x) = -\frac{1}{\lambda}\ln \left( \mathbb{E} \left[ \exp \left(-\lambda g(x+\sqrt{2}W_{T-t} \right) \right] \right),
\end{equation}
which was used to calculate the error from the true solution.

\section{Reduction of the Boussinesq Equations \label{SI:Climate}}

\begin{figure}
	\centering
	\includegraphics[width=\linewidth]{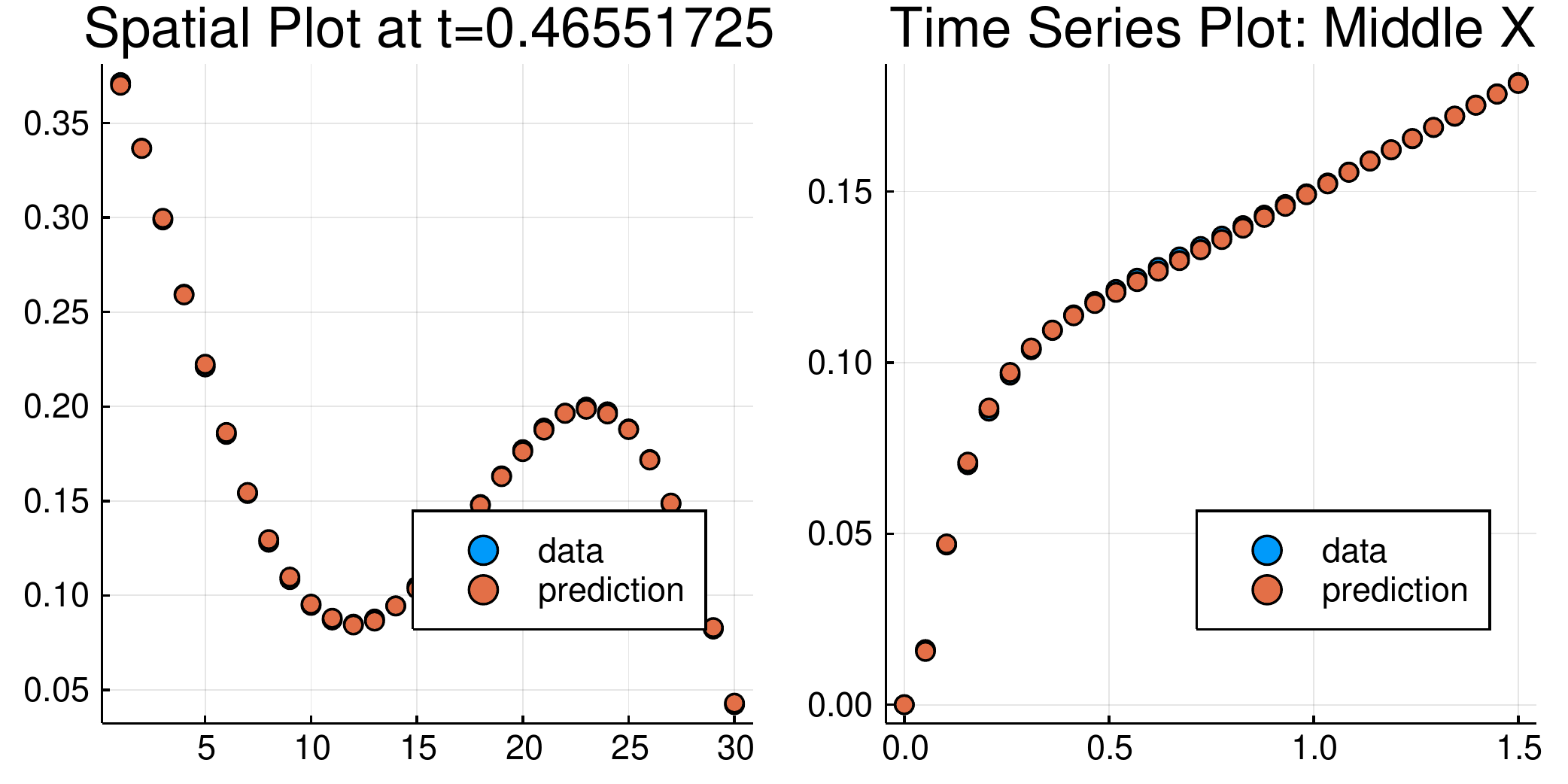}
	\caption{Reduction of the Boussinesq equations. On the left is the comparison between the training data (blue) and the trained UPDE (orange) over space at the 10th fitting time point, and on the right is the same comparison shown over time at spatial midpoint.}
	\label{SIfig:climate}
\end{figure}

As a test for the diffusion-advection equation parameterization approach, data was generated from the diffusion-advection equations using the missing function $\overline{wT} = \cos(\sin(T^3)) + \sin(\cos(T^2))$ with $N$ spatial points discretized by a finite difference method with $t \in [0,1.5]$ with Neumann zero-flux boundary conditions. A neural network with two hidden layers of size 8 and $\tanh$ activation functions was trained against 30 data points sampled from the true PDE. The UPDE was fit by using the ADAM optimizer with learning rate $10^{-2}$ for 200 iterations and then ADAM with a learning rate of $10^{-3}$ for 1000 iterations. The resulting fit is shown in \ref{SIfig:climate} which resulted in a final loss of approximately $0.007$. We note that the stabilized adjoints were required for this equation, i.e. the backsolve adjoint method was unstable and results in divergence and thus cannot be used on this type of equation. The trained neural network had a forward pass that took around 0.9 seconds.

For the benchmark against the full Bossinesq equations, we utilized Oceananigans.jl \cite{OceananigansJOSS}. It was set to utilize adaptive time stepping to maximize the time step according to the CFL condition number (capped at $\text{CFL} \le 0.3$) and matched the boundary conditions, along with setting periodic boundary conditions in the horizontal dimension. The Bossinesq simulation used $128\times128\times128$ spatial points, a larger number than the parameterization, in order to accurately resolve the mean statistics of the 3-dimensional dynamics as is commonly required in practice \cite{CUSHMANROISIN201199}. The resulting simulation took 13,737 seconds on the same computer used for the neural diffusion-advection approach, demonstrating the approximate 15,000x acceleration.

\section{Automated Derivation of Closure Relations for Viscoelastic Fluids \label{SI:Fluid}}
The full FENE-P model is:
\begin{align}\label{eq:PDE}
	\bm{\sigma} + g\left(\frac{\lambda}{f(\bm{\sigma})} \bm{\sigma} \right) &= \frac{\eta}{f(\bm{\sigma})} \bm{\dot{\gamma}},\\
	f(\bm{\sigma}) &= \frac{L^2 + \frac{\lambda (L^2 -3)}{L^2 \eta}\text{Tr}(\bm{\sigma})}{L^2-3},
\end{align}
where
$$
g(\bm{A}) = \frac{D\bm{A}}{Dt} - (\nabla \bm{u}^T)\bm{A} - \bm{A}(\nabla \bm{u}^T),
$$
is the upper convected derivative, and $L$, $\eta$, $\lambda$ are parameters \cite{Oliveira2009}. For a one dimensional strain rate, $\dot{\gamma} = \dot{\gamma}_{12} = \dot{\gamma}_{21} \neq 0$, $\dot{\gamma}_{ij} = 0$ else, the one dimensional stress required is $\sigma = \sigma_{12}$. However, $\sigma_{11}$ and $\sigma_{22}$ are both non-zero and store memory of the deformation (normal stresses). The Oldroyd-B model is the approximation:

\begin{equation}\label{eq:OldroydBG}
	G(t) = 2\eta \delta(t) + G_0 e^{-t/\tau},
\end{equation}
with the exact closure relation:
\begin{align}\label{eq:OldroydB2}
	\sigma(t) &= \eta \dot{\gamma}(t) + \phi, \\
	\frac{\mathrm{d} \phi}{\mathrm{d} t} &= G_0 \dot{\gamma} - \phi/\tau.
\end{align}

As an arbitrary nonlinear extension, train a UDE model using a single additional memory field against simulated FENE-P data with parameters $\lambda=2$, $L=2$, $\eta=4$. The UDE model is of the form,
\begin{align}
	\sigma &= U_0(\phi,\dot{\gamma}) \\
	\frac{\mathrm{d} \phi}{\mathrm{d}t} &= U_1(\phi, \dot{\gamma})
\end{align}
where $U_0,U_1$ are neural networks each with a single hidden layer containing 4 neurons. The hidden layer has a tanh activation function. The loss was taken as $\mathcal{L} = \sum_i (\sigma(t_i) - \sigma_{\text{FENE-P}}(t_i))^2$ for 100 evenly spaced time points in $t_i \in [0,2\pi]$, and the system was trained using an ADAM iterator with learning rate 0.015. The fluid is assumed to be at rest before $t=0$, making the initial stress also zero.

\end{document}